\documentclass[runningheads]{llncs}

\usepackage{eccv}

\usepackage{eccvabbrv}

\usepackage{graphicx}
\usepackage{booktabs}
\usepackage{wrapfig}
\usepackage{amssymb}
\usepackage{multirow} 
\usepackage{colortbl}
\usepackage{subcaption}
\usepackage{float}
\usepackage{multirow} 
\usepackage{longtable}
\usepackage[linesnumbered, ruled]{algorithm2e}
\SetKwRepeat{Do}{do}{while}
\usepackage{color}

\usepackage[accsupp]{axessibility}  

\usepackage[accsupp]{axessibility}  

\usepackage{hyperref}

\usepackage{orcidlink}

\begin{document}

\title{POA: Pre-training Once for Models of All Sizes} 

\titlerunning{POA}

\author{Yingying Zhang\orcidlink{0009-0006-4215-5416} \and
Xin Guo \and Jiangwei Lao \and Lei Yu \and Lixiang Ru \and Jian Wang \and Guo Ye \and Huimei He \and Jingdong Chen \and Ming Yang}

\authorrunning{Y.~Zhang et al.}

\institute{Ant Group. \\
~\email{qichu.zyy@antgroup.com}}

\maketitle

\setlength{\textfloatsep}{3pt plus 1.0pt minus 1.0pt}  
\setlength{\floatsep}{3pt plus 1.0pt minus 1.0pt}    
\setlength{\intextsep}{3pt plus 1.0pt minus 1.0pt}

\begin{abstract}
\label{sec:abs}
Large-scale self-supervised pre-training has paved the way for one foundation model to handle many different vision tasks. Most pre-training methodologies train a single model of a certain size at one time. Nevertheless, various computation or storage constraints in real-world scenarios require substantial efforts to develop a series of models with different sizes to deploy. Thus, in this study, we propose a novel tri-branch self-supervised training framework, termed as POA (\textbf{P}re-training \textbf{O}nce for \textbf{A}ll), to tackle this aforementioned issue. Our approach introduces an innovative elastic student branch into a modern self-distillation paradigm. At each pre-training step, we randomly sample a sub-network from the original student to form the elastic student and train all branches in a self-distilling fashion. Once pre-trained, POA allows the extraction of pre-trained models of diverse sizes for downstream tasks. Remarkably, the elastic student facilitates the simultaneous pre-training of multiple models with different sizes, which also acts as an additional ensemble of models of various sizes to enhance representation learning. Extensive experiments, including k-nearest neighbors, linear probing evaluation and assessments on multiple downstream tasks demonstrate the effectiveness and advantages of our POA. It achieves state-of-the-art performance using ViT, Swin Transformer and ResNet backbones, producing around a hundred models with different sizes through a single pre-training session. The code is available at: \url{https://github.com/Qichuzyy/POA}.
\keywords{Self-supervised Learning \and Pre-training Once for All}
\end{abstract}

%

\section{Introduction}
\label{sec:intro}
Learning generalizable visual representations in a large model by self-supervised learning has delivered superior performance across a wide range of visual tasks \cite{mask_rcnn, upernet, litm, img_super, img_retrieval, obj_tracking} in recent years. Nevertheless, when deployed to real-world applications, a large model has to be adapted to diverse resource limitations in terms of computation, storage, or power consumption, \etc. For example, a well-engineered AI product typically comprises a suite of models tailored for varying scenarios, such as Gemini Nano, Pro and Ultra \cite{gemini}. Given a large pre-trained model, common solutions to deploy it to multiple application scenarios with different resource constraints include additional weight pruning \cite{Yu2022WidthD, rnp, skipnet}, knowledge distillation \cite{Hinton2015DistillingTK, kds}, or even re-training a small network from scratch, which all require substantial development efforts. Consequently, this issue prompts a critical question: is it possible to pre-train once to produce multiple models with different sizes simultaneously, each delivering sufficiently good representations?

To address this challenge, we introduce a new paradigm of self-supervised learning, called POA (\textbf{P}re-training \textbf{O}nce for \textbf{A}ll). POA is built upon the prevalent teacher-student self-distillation framework \cite{dino, dinov2, ibot}, with an additional innovative elastic student branch. The elastic student branch embeds a series of sub-networks through parameter sharing, upon the observation that smaller-sized models are sub-networks of a larger-sized one for modern network structures \cite{resnet,vit,swin}. Moreover, the parameters of this branch are shared with the original, or intact student. At each pre-training step, we randomly sample a subset of parameters from the intact student to form the corresponding elastic student. Both the original intact student and the elastic student are trained to emulate the output of the teacher network. The teacher itself is continually refined via an exponential moving average (EMA) of the student’s parameters, including the sampled elastic student. The elastic student facilitates effective and efficient pre-training on different subsets of parameters, leading to the successful extraction of high-performance sub-networks from the pre-trained teacher for subsequent downstream scenarios. It also acts as a form of training regularization by enforcing the outputs to match between the teacher and various sub-networks, contributing to a stable training process. Additionally, by serving as an ensemble of different sub-networks across different pre-training steps, the elastic student improves the representation learning \cite{model_soups}.

\begin{wrapfigure}{r}{6cm}

\centering
\includegraphics[width=0.5\textwidth]{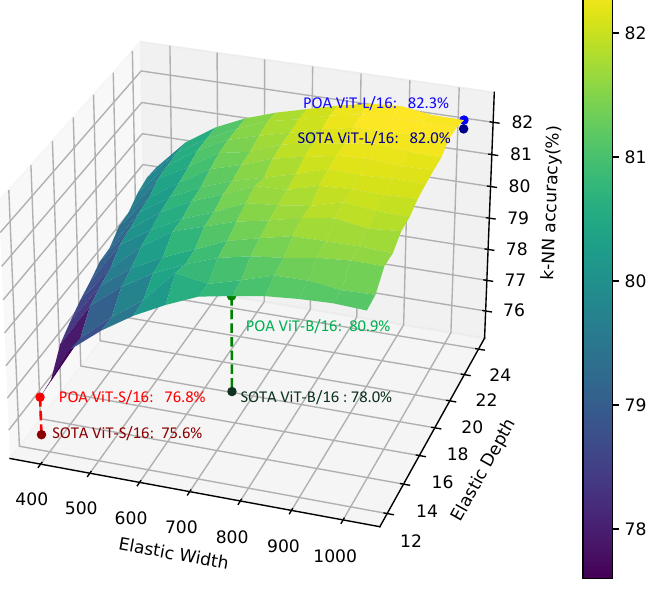}
\caption{The k-NN evaluation accuracy of 143 elastic ViTs derived from the ViT-L/16 teacher model pre-trained with POA. } 
\label{elastic_acc}

\end{wrapfigure}

To the best of our knowledge, POA represents the first self-supervised learning method capable of training multiple-sized models concurrently, each obtaining high-quality representations for different resource constraints without further pre-training. Figure \ref{elastic_acc} displays the k-nearest neighbors (k-NN) evaluation results for 143 sub-networks extracted from a ViT-L model \cite{vit} pre-trained by POA. By choosing different elastic widths and depths, which will be explained in Sec. \ref{sec:poa_1}, the pre-trained teacher model can generate a sufficient number of candidate sub-networks for the selection of the suitable model tailored for downstream applications according to the computational resources available. Notably, each sub-network is well-trained and exhibits superior performance, thanks to our meticulous design on same-view distillation, as validated in Sec. \ref{set:ablation}. In particular, the ViT-S, ViT-B and ViT-L models set new benchmarks, achieving the state-of-the-art (SOTA) results compared with those pre-trained by existing methods \cite{dinov2, ibot, ent}.

To rigorously evaluate the efficacy of our approach, we conduct extensive experiments using three widely-used backbone architectures, \ie,  ViT \cite{vit}, Swin Transformer \cite{swin}, and ResNet \cite{resnet}. Each backbone is pre-trained on ImageNet-1K dataset and assessed using k-NN and linear probing classification evaluation, as well as downstream dense prediction tasks such as object detection and semantic segmentation. Our method achieves SOTA accuracy across various model sizes with a single pre-training session. The technical contributions of this work are summarized as follows:
\begin{itemize}
    \item To the best of our knowledge, POA is the first pre-training paradigm that integrates unsupervised representation learning and once-for-all model generation within a single pre-training session. It tackles the \emph{\textbf{Pre-training-Once-for-All}} challenge which has been seldomly explored by the community but is of great practical significance for real-world deployment that usually requires a suite of models.
    \item We propose a novel and elegant component called \emph{\textbf{Elastic Student}}, featuring a range of elastic operators that enable POA to be compatible with popular backbone structures including ViT, Swin Transformer and ResNet. It provides the capability to generate models of diverse sizes. Furthermore, it serves as a model ensemble to smooth training process and improve learned representations.
    \item Through thorough assessments using k-NN, linear probing and downstream dense task evaluation, our approach exhibits superior performance over existing state-of-the-art pre-training methods across multiple metrics. Moreover, we compare our POA with Self-Supervised Distillation (SEED) \cite{seed}, a knowledge distillation method designed especially for self-supervised learning, further validating POA's effectiveness.
\end{itemize}


\section{Related Work}
\label{sec:related}
\subsection{Self-supervised Learning}
\label{sec:related_1}
Self-supervised learning (SSL) frameworks commonly fall into two categories, generative  and discriminative SSL. Most generative SSL approaches \cite{mae, simmim, spark, pcae, cmae, sdae, ss_gan, ss_gan_la, vqgan} focus on learning image representations directly in pixel space. On the other hand, discriminative SSL methods \cite{moco, simclr, swav, densecl, moby, unigrad, univip, caco} aim to learn representations by pulling those of different views of the same image closer while separating the representations of views from different images. 

Contrastive learning (CL) with the InfoNCE loss \cite{cpc} has emerged as a popular approach for discriminative SSL methods, attracting significant research interests in recent years. Although CL methods prevent the collapse of network representations through the use of negative samples, they still suffer from the dimensional collapse, where representations tend to collapse into a low-dimensional manifold. Grill \etal introduced a distillation-based asymmetric framework known as BYOL \cite{byol}, which circumvents collapse without self-labeling or contrastive loss relying on negative samples. Following this work, distillation-based frameworks have become a prevailing trend in self-supervised learning. These frameworks \cite{disco, moby, mocov2} often merge with others to enhance overall performance.
DINO \cite{dino} presented a simple self-distillation framework and has demonstrated impressive results in ViT pre-training. Subsequent works \cite{dinov2, ibot} further improved the pre-training performance via masked token or the novel clustering design. Given the substantial benefits of distillation-based methods over other SSL techniques, we have developed our POA SSL framework upon these successful methodologies.

\subsection{Dynamic Architecture.}
\label{sec:related_2}
Recently, Chen \etal proposed AutoFormer \cite{autoformer}, which trained a supernet to support the effective search of optimal sub-network under some specific parameter number constraints. On the basis of \cite{autoformer}, MaskTAS \cite{mask_tas} introduced a self-supervised transformer architecture search method. Cai \etal \cite{onceforall} trained a network that accommodates various architectural configurations to reduce the training expense. Their methodologies enable the extraction of a specialized sub-network from the trained supernet. The design of the elastic student in our POA SSL is inspired by the weight-sharing strategy employed in these neural architecture search (NAS) methods. However, our implementation differs significantly due to the distinct purpose from NAS. Specifically, NAS aims to discover the optimal architecture within certain parameter constraints, which typically involves a huge number of sub-networks (more than $10^{16}$) in the search space. Given the limitations on the number of training iterations and the network parameter capacity, it is challenging to ensure high performance across all sub-networks. After training the supernet, NAS requires a subsequent phase of searching and re-training to finalize the output model. In contrast, our POA SSL defines a sufficiently large yet compact set of sub-networks with elastic depths and widths for the purpose of pre-training models of various sizes via a single training session. Additional design on the same-view distillation guarantees that all elastic sub-networks within our framework are adequately and efficiently trained. As a result, our POA can readily extract a range of sub-networks from the teacher model without the need for extra pre-training, facilitating an easy selection of a suitable sub-network for different computational contexts. The design of elastic student is somewhat akin to the supervised learning method, Cosub\cite{cosub}. The main difference is that Cosub simply skips blocks, making only the depth elastic.

\section{POA Self-supervised Learning Framework}
\label{sec:poa}
\begin{figure}[!ht]
  \centering
  \includegraphics[height=4.2cm]{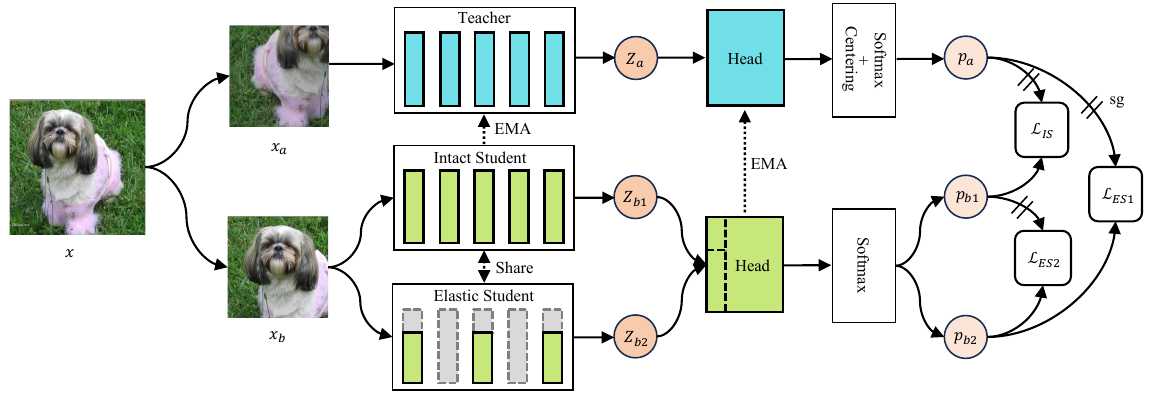}
  \caption{Overview of the POA SSL: Given an image $x$, two augmented views $x_a$ and $x_b$ are generated. These views are input into three branches: a teacher, an intact student, and an elastic student, the latter being derived from the intact student. POA optimizes distillation losses in a twofold manner: the intact and the elastic students are distilled from the teacher using the cross-view data respectively, and additionally, the elastic student is distilled from the intact student using the same-view data.}

  \label{fig:poa}
\end{figure}
Our primary goal is to pre-train models of multiple sizes via a single self-supervised pre-training session. To this end, we propose a novel SSL framework named POA, inspired by the latest progress in self-distillation techniques \cite{dino, ibot, dinov2}. The architecture of POA is illustrated in Figure \ref{fig:poa}, encompassing a teacher, an intact student, an elastic student and their corresponding heads. The teacher is updated with an EMA of the students. The elastic student is a derivative of the intact one, with both the backbones and heads' parameters shared. We leverage distillation in two aspects: both the intact and elastic students are distilled from the teacher using different views of the same image, while the elastic student additionally learns from the intact student using the identical views. The cross-view distillation works as a form of representation learning, as introduced in \cite{dino, ibot, dinov2}. Notably, in addition to the regular EMA update with only the intact student as \cite{dino, ibot, dinov2}, the elastic student provides a randomly-sampled sub-network at each pre-training step to participate in the teacher's EMA refinement. This procedure actually simulates an ensemble of multiple sub-networks, which is also proven to be beneficial in the realm of supervised learning \cite{model_soups}. The same-view distillation is a standard knowledge distillation between the intact and elastic students, promoting the quality of the elastic one. Experiments in Sec. \ref{set:ablation} validate our design in details.


\subsection{Design of Elastic Student}
\label{sec:poa_1}
The elastic student is a sub-network whose parameters are extracted from the intact student. In the context of a transformer backbone, the width refers to the dimensionality of the tokens, whereas for a convolutional backbone, the width indicates the number of channels. We denote the depth as the number of basic blocks in either a transformer or a convolutional network. Given the value of the width and depth, it yields a certain network structure. For simplicity, we focus on detailing the elastic design of the ViT in this section. For the similar elastic design of the Swin Transformer and ResNet, please refer to Appendix A.

\begin{figure}[!ht]

  \centering
  \includegraphics[height=3.8cm]{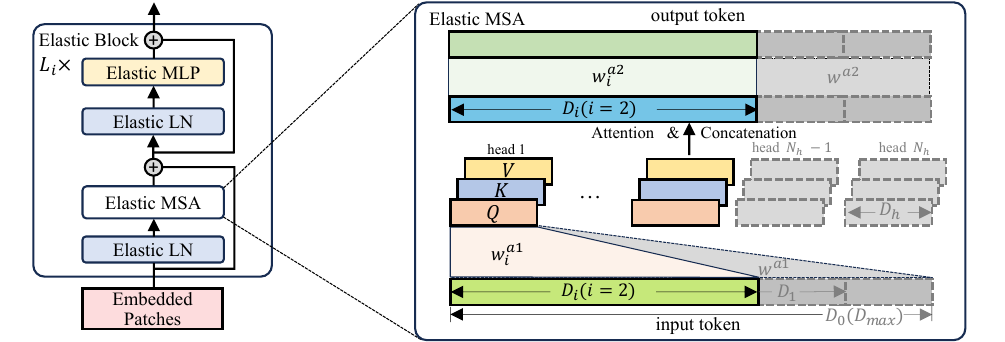}
  \caption{Illustration of the elastic MSA in an elastic ViT block. To be concise, we simply exclude the projection layers that correspond to $K$ and $V$ in each head.}

  \label{fig:elastic_student}
\end{figure}

A basic block of ViT mainly consists of a multi-head self-attention (MSA) module and a multi-layer perceptron (MLP) module. Layer Normalization (LN) \cite{layer_norm} is applied before each module, with residual connections after each module. As shown in the left part of Figure \ref{fig:elastic_student}, an elastic block refers to a stack of elastic MSA, MLP, and LN after adjusting the width in the original basic block in ViT. In our approach, the elastic student branch is constructed by assembling a specific number of these elastic blocks at each training iteration.

\paragraph{\textbf{Elastic MSA}}  
An original, or intact MSA module consists of three major components, \ie, input projection layers, an operator contains attention \& concatenation, and an output projection layer. We define a projection layer as ($w^{\ast}, b^{\ast}$), where $w^{\ast}$ is the linear transformation weight, $b^{\ast}$ denotes the corresponding bias, and $\ast$ indicates the name of the layer. As shown in the right part of Figure \ref{fig:elastic_student}, given a token dimension of $D_{max}=N_h \cdot D_h$, where $N_h$ is the number of attention heads and $D_h$ is the head dimension, an input sequence $z \in \mathbb{R}^{T \times D_{max}}$ with length $T$ is initially projected to form the queries $Q \in \mathbb{R}^{T \times D_h}$, keys $K \in \mathbb{R}^{T \times D_h}$, and values $V \in \mathbb{R}^{T \times D_h}$ by the intact MSA. To generate elastic MSA, we define M + 1 elastic widths, including $D_{max}$, spaced at intervals of $D_h$ as follows:
\begin{equation}
D_i = (N_h - i) \cdot D_h,\quad\forall i \in \{0, 1, ..., M\},\quad
M < N_h.
\label{eq:elastic_width}
\end{equation}
For each elastic width $D_i$, the weight $w^{a1}_i \in \mathbb{R}^{D_h \times D_i}$ and bias $b^{a1}_i \in \mathbb{R}^{D_h}$ that generate $Q$, $K$, and $V$ of each head are extracted from the corresponding input projection layer ($w^{a1}$, $b^{a1}$) in the intact MSA, as $w^{a1}_i = w^{a1}[:, :D_i]\cdot\alpha_i$ and $b^{a1}_i = b^{a1}$. Here, $\alpha_i$ represents the scaling factor to cope with the reduction of the input dimension, calculated as $\alpha_i = D_{max}/D_i$. As the reduction of width, the number of attention heads in the elastic MAS is naturally reduced to $N_h - i$. Similarly, for the output projection layer ($w^{a2}$, $b^{a2}$), the weight $w^{a2}_i \in \mathbb{R}^{D_i \times D_i}$ and bias $b^{a2}_i \in \mathbb{R}^{D_i}$ are extracted as:
\begin{equation}
w^{a2}_i = w^{a2}[:D_i, :D_i]\cdot\alpha_i ~~~~~ b^{a2}_i = b^{a2}[:D_i].
\label{eq:w_b_extract}
\end{equation}
\paragraph{\textbf{Elastic MLP}}
An original, or intact MLP module in the ViT block contains two projection layers. The first layer ($w^{m1}, b^{m1}$) expands the dimension of embedding by a factor of $s$, which is generally set to 4 in the ViT structure. The second layer ($w^{m2}, b^{m2}$) then projects it back to the original dimension. The parameters for both layers of the elastic MLP are extracted in a manner analogous to that described in Equation \ref{eq:w_b_extract} as follows:
\begin{equation}
\begin{aligned}
& w^{m1}_i = w^{m1}[:D_i \cdot s, :D_i]\cdot\alpha_i ~~~~~ b^{m1}_i = b^{m1}[:D_i \cdot s] \\
& w^{m2}_i = w^{m2}[:D_i, :D_i \cdot s]\cdot\alpha_i ~~~~~ b^{m2}_i = b^{m2}[:D_i].
\end{aligned}
\label{eq:mlp_extract}
\end{equation}
\paragraph{\textbf{Elastic LN}}
For elastic LN, we directly use the first $D_i$ elements of the parameter inside the original LN, akin to the bias extraction in Equation \ref{eq:w_b_extract}.

To create a sub-network with $L_i$ elastic blocks from the intact ViT comprising $L_{max}$ blocks, we introduce a set of $N + 1$ elastic depths, defined as $L_i = L_{max} - i,~~\forall i \in \{0, 1, ..., N\},~~N < L_{max}$. For a specific depth $L_i$, we select the corresponding blocks based on their block IDs at equal intervals. Each block ID $BID^{L_i}_j$ that is activated for depth $L_i$ can be formulated as:

\begin{equation}
BID^{L_i}_j = \left\lfloor \frac{(L_{max} - 1) \cdot j}{L_i - 1} \right\rfloor,\quad \forall j \in \{0, 1, ..., L_i - 1\}.
\label{eq:block_id_extraction}
\end{equation}
Consequently, by combining elastic widths and depths, we can generate a total of $(N + 1) \cdot (M + 1)$ distinct sub-networks. For instance, by setting the elastic width to 384 and the elastic depth to 12, we can directly extract a ViT-S from the intact network such as ViT-L. During each iteration of the pre-training, we randomly select one of these sub-networks to serve as the elastic student branch.

\subsection{Distillation between Views}
\label{sec:poa_2}
 POA performs distillation across its three branches accordingly. Given a pair of globally augmented views of an input image $x$, denoted as $x_a$ and $x_b$, the teacher encoder $E_{T}$ extracts the feature $Z_a = E_{T}(x_a)$ using $x_a$ as input. Simultaneously, $x_b$ is fed into both the intact student encoder $E_{IS}$ and the elastic student encoder $E_{ES}$, producing the features $Z_{b1} = E_{IS}(x_b)$ and $Z_{b2} = E_{ES}(x_b)$, respectively. The feature output from the teacher encoder, $Z_a$, is then processed by the teacher head $H_T$, followed by centering with the Sinkhorn-Knopp (SK) \cite{sk} algorithm and normalization using a temperature-scaled softmax to generate the probability $p_a$ as follows:

 \begin{equation}
\begin{aligned}
     l_a = SK(H_T(Z_a)),~l_a \in \mathbb{R}^P  ~~\quad p^i_a = \frac{\exp(l^i_a / \tau)}{\sum^{P-1}_{k=0}\exp(l^k_a / \tau)},~\forall i \in \{0, ..., P-1\},
\end{aligned}
\label{eq:prob_teacher}
\end{equation}
where $P$ is the number of prototypes and $\tau > 0$ is a temperature parameter. In a similar fashion, we compute the probabilities $p^i_{b1}$ and $p^i_{b2}$ for the intact and elastic student encoders, respectively, by processing the outputs through the student heads $H_{IS}$ and $H_{ES}$. These outputs are then passed through a temperature-scaled softmax function using a temperature parameter $\tau'$ tailored for the student. It should be noted that $H_{IS}$ and $H_{ES}$ share the identical parameters, except that the first projection layer of $H_{ES}$ is adapted similarly as in Equation \ref{eq:w_b_extract} to conform to align with the corresponding dimensionality. For simplicity, we omit the explicit expressions for $p^i_{b1}$ and $p^i_{b2}$, as they follow a similar calculation to Equation \ref{eq:prob_teacher}. For the intact student branch, we perform distillation from the teacher using the cross-view data as follows:

\begin{equation}
\mathcal{L}^g_{IS} = -p_a \log(p_{b1}).
\label{eq:distill_is}
\end{equation}
The elastic student branch plays a pivotal role in our POA framework. To ensure adequate training of this branch, we employ a dual distillation from the teacher and intact student branch. The first distillation involves the teacher model, which utilizes the cross-view data to guide the learning of representations. The second is a distillation process with the intact student model, which uses the same-view data. This same-view distillation is responsible for transferring the representations learned by the intact student to the elastic student branch. The loss functions for this dual distillation process are formulated as follows:


\begin{equation}
\begin{aligned}
\mathcal{L}^g_{ES1} = - p_a \log(p_{b2}), \quad \mathcal{L}^g_{ES2} = - p_{b1} \log(p_{b2}). 
\label{eq:distill_es}
\end{aligned}
\end{equation}
Note that in both loss functions, we sum over all prototypes to compute the cross-entropy loss between the respective probability distributions. 

\subsection{Overall Loss of POA}
\label{sec:poa_3}
Following the SSL methods such as \cite{ibot, dino, dinov2}, we employ a multi-crop strategy \cite{swav} to create various distorted views from a single image. Apart from the two global views previously mentioned, we also generate $v$ local views with lower resolution $x_{l_1}, x_{l_2}, ..., x_{l_v}$. These local views are processed by both students to promote \emph{local-to-global} correspondence. The local distillation losses for the intact and elastic students are computed as follows:

\begin{equation}
\mathcal{L}^{l}_{IS} = - \frac{1}{v}\sum^{v}_{i=1} p_a \log(p_{l_i1}),
\label{eq:local_distill_is_loss}
\end{equation}

\begin{equation}
 \mathcal{L}^{l}_{ES1} = - \frac{1}{v}\sum^{v}_{i=1}p_a \log(p_{l_i2}), \quad \mathcal{L}^{l}_{ES2} = - \frac{1}{v}\sum^{v}_{i=1}p_{l_i1} \log(p_{l_i2}),
\label{eq:local_distill_es_loss}
\end{equation}
where $p_{l_{i1}}$ and $p_{l_{i2}}$ are the probability produced by the intact and elastic student branches for the local view $l_i$, respectively. The total distillation loss of the intact and elastic student is calculated by summing them with the factor $\lambda$:
\begin{equation}
\begin{aligned}
\mathcal{L_S} &= \lambda(\mathcal{L}^g_{IS} + \mathcal{L}^{l}_{IS}) + (1-\lambda)((\mathcal{L}^g_{ES1} + \mathcal{L}^{l}_{ES1}) + (\mathcal{L}^g_{ES2} + \mathcal{L}^{l}_{ES2})) \\
&= \lambda\mathcal{L}_{IS} + (1-\lambda)(\mathcal{L}_{ES1} + \mathcal{L}_{ES2}).
\end{aligned}
\label{eq:total_distill_loss}
\end{equation}

To ensure sufficient training of each sub-network from the elastic student, we introduce multiple projection heads (MPH) positioned after the backbone network. Each projection head has exactly the same structure, except for a different number of prototypes. For each head, the distillation loss $\mathcal{L_S}_i$  for both the intact and elastic student is calculated with Equation. \ref{eq:total_distill_loss}. Finally, the overall loss function in the POA framework with $H$ projection heads is formulated as: $\mathcal{L} = \frac{1}{H} \sum^H_{i=1}\mathcal{L_S}_i$.

\section{Experiments}

\subsection{Implementation Details}
\paragraph{\textbf{Backbones.}} We have trained our POA using ViT, Swin Transformer and ResNet backbones, respectively. For the ViT, we configure the patch size to 16 and the dimension of each head in the MSA to 64. This aligns with the configurations typically used in existing SSL methods \cite{dino, ibot, mokd, ent}. For the smallest and largest elastic networks of ViT, we choose the ViT-S and ViT-L, respectively. This leads to 11 elastic widths and 13 elastic depths, yielding a total of 143 ViT sub-networks. In the case of the Swin Transformer, we set the smallest and largest elastic networks as Swin-T and Swin-B, respectively. This configuration yields a total of 39 Swin sub-networks by multiplying the number of widths and depths as $3 \times 13$. For the ResNet, we establish the smallest and largest elastic network configurations as ResNet-50 and ResNet-152. Consequently, this setup accounts for a total of 465 ResNet sub-networks, which is the product of $3 \times 155$.

\paragraph{\textbf{Pre-Training Setup.}} We pre-train all models on the ImageNet-1K dataset \cite{imagenet} without the labels. The process employs the AdamW optimizer \cite{adamw} with a batch size of 1600, which is distributed across 32 A100 GPUs when employing a ViT backbone. We adopt a learning rate schedule that begins with a linear warm-up during the first 10 epochs, reaching a base value that is scaled proportionally to the total batch size: $\text{lr} = 0.004 \times \sqrt{\text{batch size}/1024}$, in line with \cite{dinov2}. Following this warm-up period, the learning rate is decayed with a cosine schedule. Similarly, the weight decay follows a cosine schedule, starting at 0.04 and increasing to 0.4. The student network's temperature $\tau'$ is fixed at 0.1, whereas the teacher temperature $\tau$ changes with a linear warm-up from 0.04 to 0.07 over the first 30 epochs. For further details, please refer to Appendix C.

\subsection{Experiments on ImageNet-1K} 
After unsupervised pre-training, we assess the model's performance using two widely-recognized evaluation protocols in SSL domain on ImageNet-1K dataset, \ie, k-NN and linear probing. To ensure a fair comparison between SSL methods that employ different numbers of crop views for data augmentation, Zhou \etal \cite{ibot} introduced the effective training epoch as a measure to quantify the extent of a method's pre-training. We report the effective training epochs of the SSL methods for comparison. For additional evaluations including fine-tuning, semi-supervised and unsupervised learning, please refer to Appendix D.

\begin{table}[!ht]

    \small
    \caption{Comparison results of k-NN and linear probing classification accuracy (\%) on the ImageNet-1K dataset. "Param." refers to the number of parameters. "Epo." represents the number of effective training epochs following \cite{ibot}. "/16" denotes patch size of 16. "/W7" means the window size of 7. "$*$" indicates our implementation based on official codebase.  "$\dagger$" denotes reproduced results using the released code.}
    \begin{subtable}{0.32\linewidth}
    \setlength\tabcolsep{1.9pt}%
    \begin{tabular}{lccc}
    \toprule
    Method     &Epo. &  k-NN & LP\\
    \midrule
    \multicolumn{4}{l}{ResNet-50(Param. 23M)}  \\
    VICReg &2000    &-  &73.2  \\
    SwAV & 2400 & 65.7 &75.3 \\
    DINO   & 3200 & 67.5 &75.3  \\
    UniGrad & 2400   &-  &75.5 \\
    SCFS   &3200 &68.5 &75.7 \\
    ReLICv2  &4000 &- &\textbf{77.1}  \\
    \rowcolor{green!15}\textbf{POA} &0 &\textbf{73.4}   &76.9 \\
    \midrule
    \multicolumn{4}{l}{ResNet-101(Param. 41M)} \\
    ReLICv2     &4000   &-  &78.7 \\
    \rowcolor{green!15}\textbf{POA} &0  &\textbf{75.7} &\textbf{79.1}  \\
    \midrule
    \multicolumn{4}{l}{ResNet-152(Param. 56M)} \\
    ReLICv2     &4000   &-  &79.3  \\
    \rowcolor{green!15}\textbf{POA} &2400   &\textbf{76.4}  &\textbf{79.9} \\
    \bottomrule
    \end{tabular}
    \caption{ResNet backbone.}
\end{subtable}
\small
\begin{subtable}{0.32\linewidth}
    \setlength\tabcolsep{1.5pt}%
    \begin{tabular}{lccc}
    \toprule
    Method     &Epo. &  k-NN & LP\\
    \midrule
    \multicolumn{4}{l}{Swin-T/W7(Param. 28M)} \\
    SMoG    &1200   &-  &77.7  \\
    iBOT  &1200 &75.3 &78.6 \\
    DINOv2$^*$  &1200   &75.4   &78.0 \\
    EsViT  &1200    &75.7   &78.1 \\
    \rowcolor{green!15}\textbf{POA} &0  &\textbf{77.5}  &\textbf{78.9} \\
    \midrule
    \multicolumn{4}{l}{Swin-S/W7(Param. 49M)} \\
    DINOv2$^*$  &1200 &76.1 &79.8 \\
    EsViT   &1200   &77.7   &79.5 \\
    \rowcolor{green!15}\textbf{POA} &0  &\textbf{79.3}  &\textbf{81.3} \\
    \midrule
    \multicolumn{4}{l}{Swin-B/W7(Param. 87M)} \\
    DINOv2$^*$  &1200   &76.9   &80.9 \\
    EsViT &1200 &78.9   &80.4 \\
    \rowcolor{green!15}\textbf{POA} &1200   &\textbf{79.6}  &\textbf{82.0}   \\
    \bottomrule
    \end{tabular}
    \caption{Swin backbone.}
    \label{tab:det}
\end{subtable}
\small
\begin{subtable}{0.32\linewidth}
    \setlength\tabcolsep{1.8pt}%
    \begin{tabular}{lccc}
    \toprule
    Method     &Epo. &  k-NN & LP\\
    \midrule
    \multicolumn{4}{l}{ViT-S/16(Param. 21M)} \\
    DINO    &3200 &74.5 &77.0   \\
    iBOT    &3200   &75.2   &\textbf{77.9}  \\
    ENT &3200   &75.3   &77.7 \\
    \rowcolor{green!15}\textbf{POA} &0  &\textbf{76.8}  &77.6 \\
    \midrule
    \multicolumn{4}{l}{ViT-B/16(Param. 85M)} \\
    DINO    &1600   &76.1   &78.2 \\
    ENT  &1600  &77.1   &79.1 \\
    iBOT    &1600   &77.1   &79.5   \\
    \rowcolor{green!15}\textbf{POA} &0  &\textbf{80.9}  &\textbf{81.7} \\
    \midrule
    \multicolumn{4}{l}{ViT-L/16(Param. 307M)} \\
    iBOT    &1200   &78.0   &81.0   \\
    DINOv2$^\dagger$  &1200 &82.0   &83.3 \\
    \rowcolor{green!15}\textbf{POA}     &1200   &\textbf{82.3}  &\textbf{83.6}   \\
    \bottomrule
    \end{tabular}
    \caption{ViT backbone.}
    \label{tab:change}
    \end{subtable}
 \label{knn_linear}

\end{table}

\paragraph{\textbf{k-NN and Linear Probing.}} To assess the quality of pre-trained features, we employ a k-NN classifier and a linear classifier on the frozen representations. For both the k-NN and linear probing (LP) evaluation, we follow the evaluation protocols established in \cite{dino, ibot, dinov2}. The performance of our method when being trained using ResNet, Swin Transformer and ViT backbones is reported in Table \ref{knn_linear}. Our POA SSL achieves the SOTA k-NN accuracy of $\textbf{ 82.3\%}$ and LP accuracy of $\textbf{83.6\% }$ when using the ViT-L/16 backbone. By employing the sub-network extraction approach outlined in Section \ref{sec:poa_1}, we derive the sub-networks ViT-S/16 and ViT-B/16 from the teacher ViT-L/16 without any additional pre-training. Thus, the number of effective training epochs of them is 0. Notably, the extracted ViT-S/16 achieves the SOTA k-NN accuracy of $\textbf{76.8\%}$ and LP accuracy on par with the previous SOTA of $77.9\%$ reported by iBOT \cite{ibot}. Our derived ViT-B/16 model sets new benchmarks for both k-NN and LP, with accuracy of $\textbf{80.9\%}$ and $\textbf{81.7\%}$, respectively. For the models using Swin Transformer and ResNet backbones, POA also reaches SOTA performance in k-NN and LP accuracy. The only exception is the LP accuracy of ResNet-50, which is competitive with that of ReLICv2 \cite{relic_v2}, despite the latter being trained over a much longer period (4000 epochs). The superior performance achieved across a range of backbone architectures confirms the effectiveness and versatility of our method. For additional detailed comparison with \textbf{27} existing methods, please see Appendix D.1.

\subsection{Evaluation on Downstream Visual Tasks} 

\paragraph{\textbf{Object Detection and Instance Segmentation on COCO Dataset.}} For a fair comparison, we utilize the Cascade Mask R-CNN framework \cite{cascade_rcnn,mask_rcnn}, which generates both bounding boxes and instance masks, in line with previous approach \cite{ibot}, on the COCO dataset \cite{coco}. We benchmark our results against existing SSL methods that generate pre-trained ViT-S/16 and ViT-B/16 backbones. As shown in Table \ref{det_seg}, POA boosts the bounding box average precision ($\text{AP}^b$) for ViT-S/16 from 49.4 to \textbf{50.6} and the mask average precision ($\text{AP}^m$) from 42.6 to \textbf{43.8}. When applied to ViT-B/16, POA attains an $\text{AP}^b$ of \textbf{52.4} and an $\text{AP}^m$ of \textbf{45.4}, which represents a remarkable step forward over previous SOTA.

\paragraph{\textbf{Semantic Segmentation on ADE20K Dataset.}} For the semantic segmentation task, we primarily focus on two settings on the ADE20K dataset \cite{ade20k}, following \cite{ibot}. First, akin to the linear evaluation protocol in classification, we evaluate the quality of representations by keeping the patch features fixed and only fine-tuning one linear layer. This approach offers a clear comparison of representation quality. Second, we employ the UPerNet \cite{upernet} as the task head and fine-tune the entire network. As depicted in Table \ref{det_seg}, our POA significantly outperforms the supervised baseline using the ViT-S/16 backbone, achieving a substantial increase of \textbf{2.2} in mean Intersection over Union (mIoU), and surpassing iBOT by \textbf{1.3} mIoU.  With the ViT-B/16 backbone, POA exceeds the previously best-performing method, iBOT, by \textbf{0.4} mIoU when utilizing UPerNet. Furthermore, under the assessment using solely a linear head, POA obtains an impressive improvement of \textbf{2.0} mIoU over iBOT as the performance is largely determiend by the quality of the pre-trained representation. 

\begin{table}[!ht]
  \small
  \caption{Evaluation results on downstream detection and segmentation tasks. Seg.$^\dagger$ indicates the use of a linear head for semantic segmentation.}
  \label{det_seg}
  \centering
  \setlength{\tabcolsep}{3pt}
  \begin{tabular}{ccccc|ccccc}
    \toprule
    \multirow{2}{*}{Method}& \multirow{2}{*}{Arch.} & Det. & ISeg.  &Seg. & \multirow{2}{*}{Arch.} & Det. & ISeg. &Seg.$^\dagger$ &Seg. \\
    & &$\text{AP}^b$  &$\text{AP}^m$   &mIoU & &$\text{AP}^b$  &$\text{AP}^m$   &mIoU  &mIoU\\
    \midrule
    Sup. &ViT-S/16 &46.2 &40.1  &44.5 &ViT-B/16 &49.8 &43.2 &35.4 &46.6  \\
    BEiT &ViT-S/16 &- &- &-  &ViT-B/16 &50.1 &43.5 &27.4 &45.8 \\
    DINO &ViT-S/16 &- &- &-  &ViT-B/16 &50.1 &43.4 &34.5 &46.8 \\
    iBOT &ViT-S/16 &49.4 &42.6  &45.4 &ViT-B/16 &51.2 &44.2 &38.3 &50.0 \\
    \rowcolor{green!15}POA &ViT-S/16 &\textbf{50.6} &\textbf{43.8}  &\textbf{46.7}&ViT-B/16 &\textbf{52.4}  &\textbf{45.4} &\textbf{40.3} &\textbf{50.4}\\
    \bottomrule
  \end{tabular}

\end{table}

\subsection{Ablations and Discussions}
\label{set:ablation}
In this section, we conduct an empirical analysis of POA using ViT as backbone. Our investigation includes the impact of the loss functions $\mathcal{L}_{ES1}$ and $\mathcal{L}_{ES2}$, in addition with the effectiveness of multiple projection heads. Moreover, we compare our POA with knowledge distillation techniques for self-supervised learning to demonstrate the advantages of combining once-for-all model generation with self-distillation in a unified framework. For more ablation studies, we direct readers to the Appendix D.8. In addition, we further contrast our POA with three variants tailored for elastic pre-training to showcase POA's superiority. Finally, we discuss how the elastic student facilitates the pre-training.

\begin{table}[!ht]

  \caption{Contributions of each component in POA framework. We conduct assessments with the k-NN and linear probing (LP) evaluations. MPH denotes the multiple projection heads with different numbers of prototypes.}
  \label{ablation_mph}
  \small
  \centering
  \setlength{\tabcolsep}{2pt}
  \begin{tabular}{ccc|cccccc}
    \toprule
    \multirow{2}{*}{MPH}& \multirow{2}{*}{$\mathcal{L}_{ES1}$} & \multirow{2}{*}{$\mathcal{L}_{ES2}$} &   &k-NN & & &LP &  \\
    &&&ViT-S/16 &ViT-B/16 &ViT-L/16 &ViT-S/16 &ViT-B/16 &ViT-L/16\\
    \midrule
    \checkmark &\checkmark & \checkmark &\textbf{76.8} &\textbf{80.9} &\textbf{82.3} &\textbf{77.6} &\textbf{81.7} &\textbf{83.6} \\
     &\checkmark & \checkmark &76.2 &80.7&82.2 &77.3&81.6&83.4  \\
      & & \checkmark  &75.1 &80.2 &82.2 &75.8 &81.0 &83.4\\
      &  \checkmark & &72.8 &79.1 &82.1 &75.3 &80.8 & 83.3 \\
    \bottomrule
  \end{tabular}

\end{table}

\paragraph{\textbf{Importance of Each Component.}} We evaluate the contributions of the components on a ViT backbone. Table \ref{ablation_mph} shows the performance of different component combinations. First, we note that employing multiple projection heads (MPH) enhances the learned representations for each elastic sub-network, particularly for smaller ones. The design consideration behind MPH is that for each pre-training iteration, the sub-network is chosen randomly, leading to a relatively insufficient optimization. MPH introduces different sets of prototypes, which act as multiple semantic spaces for representation learning, enabling the teacher to distill various aspects of learned knowledge into the sub-network. Furthermore, we ascertain that the same-view distillation loss $\mathcal{L}_{ES2}$ is crucial for the representation quality of elastic sub-networks. Omitting $\mathcal{L}_{ES2}$ causes a significant drop in k-NN accuracy, by 3.4\% for ViT-S/16 and 1.6\% for ViT-B/16. In addition, $\mathcal{L}_{ES2}$ is more important than cross-view distillation $\mathcal{L}_{ES1}$ in terms of sub-networks' representation quality. The underlying reason is that the cross-view distillation is for the unsupervised representation learning, while the same-view distillation improves the sub-networks by distilling previously learned representations from the intact student. Table \ref{ablation_kd} confirms that distillation from already-good representations is more effective than representation learning, especially for smaller networks. It also explains why employing three different views in our POA is unnecessary.

\begin{table}[!ht]

  \caption{Comparison with knowledge distillation for self-supervised learning. The teacher name "ViT-L/16-600" denotes a teacher model (ViT-L/16) that has been pre-trained with 600 effective epochs. The student name "ViT-S/16-600" refers to a student model (ViT-S/16) that has undergone distillation from the pre-trained teacher with 600 effective epochs. The number of the left side of "$\rightarrow$" indicates the performance of the teacher model when pre-trained individually. The number of the right side denotes the performance achieved after distillation using the SEED \cite{seed}.}
  \label{ablation_kd}
  \centering
  \setlength{\tabcolsep}{1pt}
  \begin{tabular}{cccccc}
    \toprule
    Method &Teacher & Student & Total Epochs & k-NN &LP \\
    \midrule
    DINOv2  &\multicolumn{2}{c}{ViT-S/16} &1200   &\underline{72.2} &\underline{73.1}\\
    DINOv2+SEED & ViT-L/16-600 &ViT-S/16-600 &1200 &$81.3\rightarrow\underline{74.0}$&$82.4\rightarrow\underline{75.2}$\\
    DINOv2+SEED & ViT-L/16-1200 &ViT-S/16-1200 &2400 &$82.0\rightarrow\underline{75.5}$ &$83.3\rightarrow\underline{76.2}$\\
    POA  &\multicolumn{2}{c}{ViT-S/16} &1200 &\underline{\textbf{76.8}} &\underline{\textbf{77.6}}\\
    \midrule
    DINOv2  &\multicolumn{2}{c}{ViT-B/16} &1200   &\underline{77.4} &\underline{78.5}\\
   DINOv2+SEED &ViT-L/16-600 &ViT-B/16-600 &1200 &$81.3\rightarrow\underline{78.8}$ &$82.4\rightarrow\underline{80.0}$\\
   DINOv2+SEED &ViT-L/16-1200 &ViT-B/16-1200 &2400 &$82.0\rightarrow\underline{79.7}$ &$83.3\rightarrow\underline{80.9}$\\
    POA  &\multicolumn{2}{c}{ViT-B/16} &1200 &\underline{\textbf{80.9}} &\underline{\textbf{81.7}}\\
    \bottomrule
  \end{tabular}

\end{table}

\paragraph{\textbf{Comparison with Knowledge Distillation.}} Knowledge distillation (KD) is a proven strategy for enhancing small networks by leveraging the knowledge of a well-trained, larger network. We compare our method with a self-supervised KD method SEED \cite{seed} that employs a pre-trained network from the previous SOTA DINOv2 as the teacher for distilling knowledge into smaller networks. We observe that SEED significantly boosts the performance of ViT-S/16 and ViT-B/16 when a pre-trained ViT-L/16 serves as the teacher. Specifically, with the same number of training epochs, SEED yields k-NN accuracy improvements of 1.8\% and 1.4\% for ViT-S/16 and ViT-B/16 respectively, compared with a learn-from-scratch setup. Our POA further outperforms SEED, delivering superior k-NN accuracy gains of \textbf{2.8\%} (76.8\% vs. 74.0\%) and \textbf{2.1\%} (80.9\% vs. 78.8\%) on ViT-S/16 and ViT-B/16. Moreover, the ViT-S/16 and ViT-B/16 derived directly from the pre-trained teacher from our POA, without additional pre-training, perform better than those enhanced by SEED, despite the latter undergoing two times of training epochs. These results demonstrate the effectiveness of our unified design, which integrates pre-training with once-for-all model generation. Additional comparison on training cost can be found in Appendix D.11.

\begin{figure}[!ht]

  \centering
  \includegraphics[height=3.1cm]{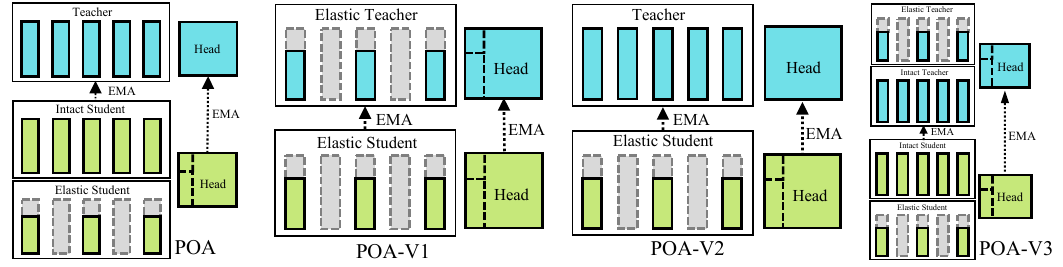}
  \caption{Illustration of different variants of POA.}
  \label{fig:POA_variants}
\end{figure}

\paragraph{\textbf{Alternative Designs to Elastic Pre-training.}} Upon the proposed POA, we investigate three alternatives for concurrently pre-training multi-sized models within a single pre-training session. As illustrated in Figure \ref{fig:POA_variants}, the first variant, POA-V1, eliminates the intact student and modifies the teacher to be elastic which aligns with the architecture of the elastic student. The second variant, POA-V2, only removes the intact student. However, in this model, the intact teacher is directly updated via an EMA of the elastic students. These two variants bear resemblance to current teacher-student self-distillation paradigm, with the primary differences in the structure of the teacher or student network. For the POA-V3, we introduce an additional elastic teacher branch that mirrors the architecture of the elastic student, enabling further cross-view distillation. For more details of the three variants, please refer to the Appendix D.9.

We compare the k-NN and linear probing accuracy of each variant in Table. \ref{ablation_cmp_poa_variant}. The results indicate that the performance drops dramatically for the variants with two branches. This decline can be attributed to two reasons: 1) in each training iteration, only a subset of the student's parameters are updated, leading to relatively insufficient overall pre-training; and 2) each elastic network undergoes training exclusively with the cross-view distillation, missing out the standard distillation guidance from a larger network, which our ablation study suggests is crucial for the efficacy of elastic networks. Although the POA-V3, featuring an additional elastic teacher branch, performs slightly better than our POA, it does so at the expense of increased model complexity and computational costs, making it a less appealing scheme than POA.

\begin{table}[!ht] 

  \caption{k-NN and linear probing (LP) comparison of different variants of POA.}

  \label{ablation_cmp_poa_variant}
  \centering
  \setlength{\tabcolsep}{5pt}
  \begin{tabular}{ccccccc}
    \toprule
    \multirow{2}{*}{Method} &   &k-NN & & &LP &  \\
    &ViT-S/16 &ViT-B/16 &ViT-L/16 &ViT-S/16 &ViT-B/16 &ViT-L/16\\
    \midrule
    POA &76.8 &80.9 &\textbf{82.3} &77.6 &81.7 &\textbf{83.6} \\
    POA-V1 &73.7 &78.5 &79.2 &74.5 &79.4 &80.9 \\
    POA-V2 &74.3 &79.1 &80.9 &75.1 &80.0 &82.3 \\
    POA-V3 &\textbf{76.9} &\textbf{81.0} &\textbf{82.3} &\textbf{77.8} &\textbf{81.8} &\textbf{83.6} \\
    \bottomrule
  \end{tabular}

\end{table}

\paragraph{\textbf{How Does Elastic Student Facilitate Pre-training?}} The elastic student branch enables the derivation of diverse-sized models directly from a pre-trained teacher, while simultaneously enhancing the learned representation. The elastic branch plays a dual role. First, it acts as a training regularization to stabilize the training progress. In our preliminary experiments, we observe that pre-training loss of a ResNet backbone easily diverges to NaN without the elastic branch. Conversely, the inclusion of the elastic student yields a highly stable pre-training. Furthermore, the elastic student provides hundreds of sub-network candidates that will be assembled to the teacher during pre-training. Unlike existing self-distillation methods, the teacher in the POA SSL integrates a series of sub-networks through an EMA update. Wortsman \etal \cite{model_soups} have shown that averaging the weights of multiple models typically improves accuracy and robustness. Similarly, the integration of different sub-networks for the teacher is effective in improving the representation quality. At the same time, an improved teacher promotes the learning of students in return, fostering a positive feedback loop for pre-training. For the visualization of representations, please refer to Appendix E.2. 

\section{Conclusion}
In this study, we tackle the challenge of efficiently and effectively pre-training models of various sizes within a single self-supervised learning session, facilitating model deployment given different resource constraints. We propose a novel self-supervised learning paradigm, termed POA, which features an integrated design combining self-distillation and once-for-all model generation. It allows for the simultaneous pre-training of models of multiple sizes through an elastic branch design. POA enables the direct generation of varied-sized models from a pre-trained teacher, which are ready for downstream tasks without additional pre-training. This advantage significantly improves the deployment flexibility and facilitates our pre-trained model to achieve SOTA results across various vision tasks. Looking forward, we plan to extend POA to Multimodal Large Language Models, tapping into its vast potential for real-world AI product deployment.

\appendix
\section{Design of Elastic Student}
\subsection{Elastic Swin Transformer}
The Swin Transformer's basic block \cite{swin} closely resembles that of the ViT, and we apply the same parameter extraction method to its Shifted Window based Self-Attention (SW-MSA) and MLP modules. In the Swin Transformer architecture, the final linear layer of the MLP in the last block of each stage quadruples the dimension of the tokens. Subsequently, a Patch Merging (PM) layer is employed to halve the token dimension. We adjust the parameters of these linear layers in the MLP and PM to match their respective expansion and reduction ratios as follows:
\begin{equation}
\begin{aligned}
& w^{mlp}_i = w^{mlp}[:D^s_i \cdot 4, :D^s_i]\cdot \alpha_i, ~~
& w^{pm}_i = w^{pm}[:D^s_i \cdot 2, :D^s_i \cdot 4]\cdot \alpha_i,
\end{aligned}
\end{equation}
where $D^s_i$ represents the $i$-th elastic width of stage $s$, $w^{mlp}$ are the weight of the last linear layer of the MLP, and $w^{pm}$ are the weight of the PM module. For the Swin Transformer, we apply elastic depth exclusively to the third stage, where the number of blocks is larger. The approach for selecting the activated block IDs follows the same method as used for the ViT, ensuring consistency across different transformer architectures in the process of creating sub-networks with varying depths.

\subsection{Elastic ResNet}
ResNet is composed of many basic units known as bottleneck building blocks \cite{resnet}. In ResNet, we primarily focus on constructing sub-networks with varying numbers of blocks. We make the feature dimension(channel) of the middle layer in each building block elastic, while keeping the output dimension from each block unchanged. The weights of the three convolutional layers within a block are extracted as follows:
\begin{equation}
\begin{aligned}
w^{1}_i = w^{1}[:D^s_i, :, :, :],~
w^{2}_i = w^{2}[:D^s_i, :D^s_i, :, :]\cdot \alpha_i,~
w^{3}_i = w^{3}[:, :D^s_i, :, :]\cdot \alpha_i,
\end{aligned}
\end{equation}
where $D^s_i$ denotes the number of mid-layer channel for stage $s$. We implement elastic depth in both the second and third stages of ResNet. The method for selecting block IDs in each stage is consistent with that used for the ViT and Swin Transformer.

\section{Pseudocode}
\vspace{-10pt}
\begin{algorithm}[H]
  \LinesNotNumbered
  \caption{POA PyTorch-like Pseudocode without multi-crop augmentation and MPH.}
  \SetKwInOut{Input}{Input}\SetKwInOut{Initialize}{Initialize}
  \Input{
  $g_{IS}, g_{T}$\quad\textcolor[RGB]{100,180,255}{\tcp*[h]{intact student and teacher network}}\\
  $M, N$\quad\textcolor[RGB]{100,180,255}{\tcp*[h]{number of elastic width and depths}}\\
  $D_{max}, L_{max}$\quad\textcolor[RGB]{100,180,255}{\tcp*[h]{max width, max depth}}\\
  $D_{h}$\quad\textcolor[RGB]{100,180,255}{\tcp*[h]{head dimension}}\\
  $\tau_s, \tau_t, \mu$\quad\textcolor[RGB]{100,180,255}{\tcp*[h]{temperatures and momentum for EMA}}\\
  $\lambda, \gamma$\quad\textcolor[RGB]{100,180,255}{\tcp*[h]{loss weight}}\\
  }
  \Initialize{
  $g_{T}$.params = $g_{IS}$.params, \quad
  cand\_ids = [[$i, j$] for $i$ in range(M + 1) for j in range(N + 1)],\quad idx=0
  }
   \SetKwProg{Fn}{def}{:}{}
   \SetKwFunction{ExtractElastic}{ExtractElastic}
   \SetKwFunction{H}{H}
   \For{
   $x$  \rm{in dataloader}
   } {
   $x_a$, $x_b$=augment(x), augment(x)\quad\textcolor[RGB]{100,180,255}{\tcp*[h]{random views}}\\
   $g_{ES}$, idx = \ExtractElastic{\rm{cand\_ids, idx, $D_{max}$, $D_{h}$, $L_{max}$, $g_{IS}$}}\\
   $f_a = g_T(x_a)$, \quad
   $f_{b1} = g_{IS}(x_b)$, \quad
   $f_{b2} = g_{ES}(x_b)$ \\
   $\mathcal{L}_{IS}$ = \H{$f_{b1}, f_a, \tau_s, \tau_t$} \\
   $\mathcal{L}_{ES}$ = $\mathcal{L}_{ES1} + \mathcal{L}_{ES2}$ = \H{$f_{b2}, f_a, \tau_s, \tau_t$} + \H{$f_{b2}, f_{b1}, \tau_s, \tau_s$, \rm{False}} \\
   $\mathcal{L} = \lambda \mathcal{L}_{IS} + (1 - \lambda)\mathcal{L}_{ES} + \gamma \mathcal{L}_{koleo} $ \quad\textcolor[RGB]{100,180,255}{\tcp*[h]{total loss with Koleo regularization}}\\
   $\mathcal{L}$.backward(),\quad
   update($g_{IS}$) \quad\textcolor[RGB]{100,180,255}{\tcp*[h]{Note that $g_{ES}$ and $g_{IS}$ share parameters, and the gradient from $g_{ES}$ is accumulated onto the gradient of $g_{IS}$.}}\\
   $g_T.\rm{params}$ $= \mu \cdot g_T.\rm{params} + (1 - \mu) \cdot $$g_{IS}.\rm{params}$\quad\textcolor[RGB]{100,180,255}{\tcp*[h]{update teacher with momentum EMA}}\\
   }
     ~~\\
    \SetAlgoNoEnd 
    \Fn{\ExtractElastic{\rm{cand\_ids, idx, $D_{max}$, $D_{h}$, $L_{max}$, $g_{IS}$}}}{
         \If{$\rm{idx} == \rm{len(cand\_ids)} - 1$}{
            \rm{random.shuffle}(\rm{cand\_ids})\\
            $\rm{idx} = 0$
        }
        $i, j =  \text{cand\_ids}[\text{idx}] $\\
        $D_i = D_{max} - i\cdot D_h$ \\
        $L_j = L_{max} - j$ \\
        $g_{ES} = \text{Net}(g_{IS}, D_i, L_j)$~\textcolor[RGB]{100,180,255}{\tcp*[h]{sub-network extracted from $g_{IS}$ with width $D_i$ and depth $L_j$}}\\
        $\rm{idx} += 1$ \\
        \KwRet{$g_{ES}, \rm{idx}$}
    }
    ~~\\
    ~~\\
    \Fn{\H{$s, t, \tau_s, \tau_t, \rm{centering=True}$}}{
     $t = t.\rm{detach}()$ \quad\textcolor[RGB]{100,180,255}{\tcp*[h]{stop gradient}}\\
     $s = \text{softmax}(s / \tau_s, \text{dim}=1)$\\
     \If{\rm{centering}}{
        $t = \text{SK}(t)$\quad\textcolor[RGB]{100,180,255}{\tcp*[h]{SK centering}}\\
    }
     $t = \text{softmax}(t / \tau_t, \text{dim}=1)$\\
     \KwRet{$-(t \cdot \log(s)).\rm{sum(dim=-1)}$}
    }
    
\end{algorithm}

\section{Implementation Details}

\subsection{Multiple Projection Heads} 
Following \cite{dino, ibot, dinov2}, we employ 3-layer MLP with L2-normalized bottlenecks to serve as projection heads. To ensure effective training of each network within these elastic frameworks, we introduce multiple projection heads, each with a varying number of prototypes, positioned subsequent to the backbone network. Our MPH design is distinct from the multiple heads utilized in the ENT \cite{ent}. In the ENT, each head contains an identical number of prototypes and employs an averaging of cross-entropy loss which is weighted by the predictive entropies of each head, to ensemble the learning of each head. In contrast, our MPH design features heads with varying numbers of prototypes, which acts as multiple semantic spaces for representation. This design is intended to improve the distillation process between the intact network and the elastic network. In our experiments, we designate the output dimensions for these projection heads as: $K_1=8192, K_2=16384, K_3=32768, K_4=65536$. 

\subsection{KoLeo Regularizer} 
To mitigate the issue of feature collapse, we incorporate the KoLeo regularizer into our training process for the global views, as described in \cite{dinov2}. The regularizer's loss function is expressed as:
\begin{equation}
\begin{aligned}
& \text{KoLeo}(z) = -\frac{1}{B}\sum_{i=1}^{B}\log(d_{B, i}), \quad d_{B, i}=\min_{j\neq i} ||z_i - z_j||, \\
& \mathcal{L}_{koleo} = \text{KoLeo}(\frac{Z_{b1}}{||Z_{b1}||}) + \text{KoLeo}(\frac{Z_{b2}}{||Z_{b2}||}),
\end{aligned}
\label{eq:koleo_loss}
\end{equation}
where $B$ is the batch size, and $d_{B, i}$ represents the Euclidean distance between the $i$-th feature $z_i$ and its nearest feature $z_j$ within the batch. The KoLeo loss is scaled with a modest loss weight $\gamma$ set to 0.1.

\subsection{Masked Patch Tokens Prediction} 
iBOT \cite{ibot} has demonstrated the effectiveness of predicting the masked patch tokens of student networks according to the tokens of teacher network. We also incorporate this mechanism into the training of our POA SSL. In our experiments, we observed that early implementation of masked patch token prediction can result in unstable training. To address this issue, we delay the activation of the masked patch token prediction until the model has completed 30 epochs of training.

\subsection{Probabilistic Sampling for Elastic Student} 
In POA, there are an array of candidate elastic networks that vary in width and depth, each offering a different level of diversity when compared to the intact student model. For example, considering the intact student structure as ViT-L characterized by a width of 1024 and depth of 24, an elastic network such as ViT-S, with a width of 384 and depth of 12, exhibits a higher degree of diversity relative to the intact student than an elastic network with dimensions closer to the intact student, such as a width of 960 and depth of 23. Intuitively, the elastic networks with greater diversity should be sampled more frequently to ensure sufficient training. To facilitate this, we implement a probabilistic sampling method influenced by the width and depth of the elastic networks in our experiments. For $N + 1$ available widths and $M + 1$ available depths, we calculate the sampling probability for the $i$-th width and $j$-th depth elastic network as follows:
\begin{equation}
p_{i,j} = \frac{((S_w - 1) \cdot \frac{N - i}{N} + 1) \cdot ((S_h - 1) \cdot \frac{M - j}{M} + 1)}{\sum_{k=0,l=0}^{N,M} ((S_w - 1) \cdot \frac{N - k}{N} + 1) \cdot ((S_h - 1) \cdot \frac{M - l}{M} + 1)}.
\end{equation}
It is important to note that when $i = 0$ and $j = 0$, the elastic network is at its smallest width and depth, and the sampling probability $p_{i,j} $ achieves the largest value. In our experiment, we set: $S_w = S_d = 3$.

\subsection{Data Augmentation Setting}
We adopt the same data augmentation techniques as DINOv2 \cite{dinov2}, which include color jittering, Gaussian blur, solarization, flipping, and multi-crop strategies as described in \cite{swav}. The specific parameters for these augmentations are detailed in Table \ref{aug_param}.
\begin{table}[!h]
\centering
\setlength{\tabcolsep}{7.0pt}
  \caption{Hyper-parameters of different data augmentations. The parameters prob\_g1, prob\_g2, and prob\_l refer to the activation probabilities for the first global crop, the second global crop, and the local crops, respectively. The parameter min\_gcs represents the minimum global crop scale, while max\_gcs indicates the maximum global crop scale. Similarly, min\_lcs and max\_lcs specifies the minimum and maximum local crop scale, respectively.}
  \begin{tabular}{cccccc}
    \toprule
    Color jittering & Gaussian blur & Solarization &  Multi-crop & Flipping \\
    \midrule
    brightness: 0.4   & radius\_min: 0.1 &thresh: 128 &  global crops: 2 & direction: \\
    contrast: 0.4     & radius\_max: 2.0 &prob\_g1: 0.0 &  local crops: 8 & horizontal \\
    saturation: 0.2   & prob\_g1: 1.0  &prob\_g2: 0.2 &  global size: 224 &prob: 0.5  \\
    hue: 0.1 &  prob\_g2: 0.1 & prob\_l: 0.0 & local size: 96  &  \\
    prob: 0.8 &prob\_l: 0.5  & &min\_gcs: 0.32 &  \\
    &  & &max\_gcs: 1.0  &  \\
    &  & &min\_lcs: 0.05  &  \\
    &  & &max\_lcs: 0.32  &  \\
    \bottomrule
  \end{tabular}
  \label{aug_param}
\end{table}

\subsection{Hyper-parameters of POA} 
We provide following hyper-parameters setting in our method:
\begin{itemize}
    \item \textbf{projection heads:} bottleneck dim: 256~~~~hidden layer dim: 2048
    \item \textbf{drop path rate:} ViT: 0.2~~~~Swin: 0.2~~~~ResNet: 0.0
    \item \textbf{loss weights:} $\lambda=0.8 ~~~\gamma=0.1$
\end{itemize}

\subsection{Optimizing Setting} 
In our training, we utilize the AdamW optimizer with parameters $\beta_1=0.9$ and $\beta_2=0.999$. The total training batch sizes for Vit, Swin, and ResNet are 1600, 2048, and 1280 respectively. We apply a learning rate decay from top to bottom across the network blocks, scaling down by a factor of 0.9. For transformer-based backbones, the patch embedding module's learning rate is further reduced by a factor of 0.2. Within each projection head, we keep the parameters of the final layer fixed during the initial epoch of training. Additionally, to maintain stable training, the gradient is clipped at an L2 norm of 1.5 for all parameters.  The momentum in EMA updating for teacher network is initialized as 0.992 and decay to 0.9999 with cosine schedule.

\section{Experiments}
\subsection{k-NN and Linear Probing Evaluation on ImageNet-1K Dataset}
We provide a more detailed comparison of k-NN and linear probing evaluations against existing methods in Table \ref{all_knn_lp}. 
\begin{center}
\setlength{\tabcolsep}{3.5pt}
\begin{longtable}{lcccccccc}
    \caption{Comprehensive comparison of k-NN and linear probing (LP) accuracy (\%) on 
     the ImageNet-1K dataset. "Param." indicates the quantity of parameters within the backbone network, measured in megabytes. "Epoch" refers to the adjusted number of effective training epochs, corrected for the number of views processed by the models as described in \cite{ibot}. "$*$" denotes our implementation based on official code. "$\dagger$" denotes results reproduced using the official code.}\\
    \toprule
    Method & Publication  & Arch.  & Param. &Epoch &  k-NN & LP\\
    \midrule
    \multirow{2}{*}{SwAV\cite{swav}} &\multirow{2}{*}{NeurIPS 20} & RN-50& 23 & 2400 & 65.7 &75.3 \\
    & &RN-200 & 250 &2000 &73.9 &79.6 \\
    BYOL\cite{byol}  &NeurIPS 20 & RN-50 & 23 & 2000 & 64.8 &74.4 \\
    MoCov3\cite{moco_v3} &ICCV 21  & RN-50 & 23 & 1600 & - & 74.6  \\
    DINO\cite{dino}  &ICCV 21 & RN-50 & 23 & 3200 & 67.5 &75.3  \\
    UniGrad\cite{unigrad} & CVPR 22 & RN-50 & 23 & 2400  &-  &75.5 \\
    \multirow{3}{*}{ReLICv2\cite{relic_v2}} &\multirow{3}{*}{arXiv 22 } & RN-50 & 23 &4000 &- &\textbf{77.1}  \\
    & & RN-101 &41  &4000   &-  &78.7 \\
    & & RN-152 &56  &4000   &-  &79.3  \\
    Univip\cite{univip} &CVPR 22 &RN-50 &23 &1200   &-  &74.2 \\
    Caco\cite{caco} &arXiv 22 &RN-50    &23 &3200 &- &75.7  \\
    SMoG\cite{smog}&ECCV 22 &RN-50  &23 &1200   &-  &76.4  \\
    VICReg\cite{vicreg} &ICLR 22 & RN-50 &23 &2000  &-  &73.2  \\
    HCSC\cite{hcsc} &CVPR 22  &RN-50 &23 &800 &66.6 &73.3  \\
    SDMP-MoCov3\cite{sdmp}  &CVPR 22  &RN-50 &23 &600 &- &73.5  \\
    SimSiam+GSG\cite{gsg}  &NeurIPS 23 & RN-50 & 23 & 400 & 58.4 &69.4 \\
    BYOL+GSG\cite{gsg}  &NeurIPS 23 & RN-50 & 23 & 400 & 62.2 &71.1 \\
    GroCo\cite{groco} &ICCV 23 & RN-50 & 23 & 400 & 64.8 &71.3 \\
    MOKD\cite{mokd}  &CVPR 23 & RN-50 & 23 & 400 & 70.6 &75.6  \\
    SCFS\cite{scfs} &CVPR 23  & RN-50 & 23 &3200 &68.5 &75.7 \\
    BYOL+LDReg\cite{LDReg} &ICLR 24 & RN-50 & 23 & 400 & - & 68.5 \\
    AUC-CL\cite{auc_cl} &ICLR 24 & RN-50 & 23 & 1400 & - & 73.5 \\
    SimCLR+WNW\cite{wnw}&ICLR 24 & RN-50 & 23 & 1600 & - & 66.3 \\
    SimSiam+WNW\cite{wnw}&ICLR 24 & RN-50 & 23 & 1600 & - & 71.3 \\
    \rowcolor{green!15}&&RN-50 & 23 &0 &\textbf{73.4}   &76.9 \\
    \rowcolor{green!15}&&RN-101 &41 &0  &\textbf{75.7} &\textbf{79.1}  \\
    \rowcolor{green!15}\multirow{-3}{*}{\textbf{POA(Ours)}} &&RN-152 &56 &2400  &\textbf{76.4}  &\textbf{79.9} \\

    \midrule
    MoBY\cite{moby} &arXiv 21 & Swin-T/W7   &28 &600    &-  &75.0 \\
    \multirow{2}{*}{iBOT\cite{ibot}} & \multirow{2}{*}{ICLR 22} & Swin-T/W7 &28 &1200 &75.3 &78.6 \\
    &&Swin-T/W14    &28 &1200   &76.2   &79.3   \\
    SMoG\cite{smog}&ECCV 22 &Swin-T/W7  &28 &1200   &-  &77.7  \\
    \multirow{6}{*}{EsViT\cite{esvit}} &\multirow{6}{*}{ICLR 22} &Swin-T/W7 &28 &1200   &75.7   &78.1 \\
    &&Swin-S/W7 &49 &1200   &77.7   &79.5 \\
    &&Swin-B/W7 &87 &1200   &78.9   &80.4 \\
    &&Swin-T/W14    &28 &1200   &77.0   &78.7 \\
    &&Swin-S/W14    &49 &1200   &79.1   &80.8 \\
    &&Swin-B/W14    &87 &1200   &79.3   &81.3 \\
    \multirow{3}{*}{DINOv2$^*$\cite{dinov2}} &\multirow{3}{*}{TMLR 24} &Swin-T/W7   &28 &1200   &75.4   &78.0 \\
    &&Swin-S/W7 & 49    &1200   &76.1   &79.8 \\
    &&Swin-B/W7 & 87    &1200   &76.9   &80.9 \\
    \rowcolor{green!15}&&Swin-T/W7 &28 &0   &\textbf{77.5}  &\textbf{78.9} \\
    \rowcolor{green!15}&&Swin-S/W7  &49 &0  &\textbf{79.3}  &\textbf{81.3} \\
    \rowcolor{green!15}\multirow{-3}{*}{\textbf{POA(Ours)}} &&Swin-B/W7 & 87    &1200   &\textbf{79.6}  &\textbf{82.0}   \\

    \midrule
    SwAV\cite{swav} &NeurIPS 20 & ViT-S/16 & 21 &2400 &66.3 &73.5 \\
    \multirow{3}{*}{MoCov3\cite{moco_v3}} &\multirow{3}{*}{ICCV 21}   & ViT-S/16    &21 &1200   &-  &73.4 \\
     & &ViT-B/16    &85 &1200   &-  &76.7 \\
     & &ViT-L/16    &307 &1200  &-  &77.6 \\
     \multirow{2}{*}{DINO\cite{dino}}  &\multirow{2}{*}{ICCV 21} & ViT-S/16 &21 &3200 &74.5 &77.0   \\
     & &ViT-B/16    &85 &1600   &76.1   &78.2 \\
    \multirow{3}{*}{iBOT\cite{ibot}} &\multirow{3}{*}{ICLR 22}   &ViT-S/16 &21  &3200   &75.2   &77.9   \\
    &&ViT-B/16  &85 &1600   &77.1   &79.5   \\
    &&ViT-L/16  &307    &1200   &78.0   &81.0   \\
     \multirow{2}{*}{SDMP-MoCov3\cite{sdmp}} &\multirow{2}{*}{CVPR 22} &ViT-S/16    &21 &600    &-  &73.8 \\
     &&ViT-B/16 &85 &600 &- &77.2\\
    SDMP-DINO\cite{sdmp} &CVPR 22 & ViT-S/16    &21 &1200 &-    &76.4\\
    \multirow{3}{*}{Mugs\cite{mugs}} &\multirow{3}{*}{arXiv 22}   &ViT-S/16 &21 &3200   &75.6   &\textbf{78.9}  \\
    &&ViT-B/16  &85 &1600   &78.0   &80.6   \\
    &&ViT-L/16  &307    &1000   &80.3   &82.1   \\
    \multirow{3}{*}{MSN\cite{msn}} &\multirow{3}{*}{ECCV 22}   &ViT-S/16    &21 &800    &73.3   &76.6   \\
    &&ViT-B/16  &85 &400    &74.7 &78.1 \\
    &&ViT-B/8 &85   &300    &75.7 &80.3 \\
    \multirow{2}{*}{MOKD\cite{mokd}} &\multirow{2}{*}{CVPR 23}   &ViT-S/16  &21 &800    &73.1   &76.3   \\
    &&ViT-B/16  &85 &400    &76.0   &78.4   \\
    \multirow{2}{*}{I-JEPA\cite{mokd}} &\multirow{2}{*}{CVPR 23}   &ViT-B/16    &85 &600    &-  &72.9 \\
    &&ViT-L/16  &307    &600    &-  &77.5 \\
    SiameseIM\cite{siameseIM}  &CVPR 23 &ViT-B/16   &85 &1600   &-  &78.0\\
    \multirow{2}{*}{ENT-DINO\cite{ent}}  &\multirow{2}{*}{ICLR 23}  &ViT-S/16   &21 &3200   &75.2   &77.4 \\
    &&ViT-B/16  &85 &1600   &77.1   &79.1 \\
    \multirow{3}{*}{ENT-MSN\cite{ent}}  &\multirow{3}{*}{ICLR 23}  &ViT-S/16    &21 &800    &75.2   &77.4   \\
    &&ViT-B/16  &85 &400    &77.2 &78.9 \\
    &&ViT-B/8 &85   &300    &78.9 &80.8 \\
    SimCLR+LDReg\cite{LDReg} &ICLR 24 &ViT-B/16  &85 &800 & - & 73.0 \\
    \multirow{3}{*}{DINOv2$^\dagger$\cite{dinov2}} &\multirow{3}{*}{TMLR 24} &ViT-S/16  &21 &1200   &72.2   &73.1 \\
    &&ViT-B/16 & 85 &1200   &77.4   &78.5 \\
    &&ViT-L/16 & 307    &1200   &82.0   &83.3 \\
    AUC-CL\cite{auc_cl} &ICLR 24 & ViT-S/16 & 21 & 1400 &70.7 &73.7 \\
    \rowcolor{green!15}&&ViT-S/16 &21 &0    &\textbf{76.8}  &77.6 \\
    \rowcolor{green!15}&&ViT-B/16  &85 &0   &\textbf{80.9}  &\textbf{81.7} \\
    \rowcolor{green!15}\multirow{-3}{*}{\textbf{POA(Ours)}} &&ViT-L/16 & 307    &1200   &\textbf{82.3}  &\textbf{83.6}   \\
    \bottomrule
\label{all_knn_lp} 
\end{longtable}
\end{center}

\subsection{Fine-Tuning Evaluation}
Due to the fine-tuning process adjusting the pretrained parameters of the backbone network, the differences between pretrained features are diminished. This may result in comparisons that may not fully reflect the distinct qualities of leaned representation in each method. Consequently, only a handful of studies report this metric. In our experiments, we conduct fine-tuning on the ImageNet-1K dataset and draw comparisons to self-supervised methods utilizing ViT backbone.  We adhered to the fine-tuning methodology delineated in \cite{beit,ibot}, which incorporates layer-wise learning rate decay, weight decay, and the AdamW optimizer. The training durations for ViT and Swin variants are set at 200, 100, and 50 epochs for the large, base, and small models, respectively. Due to differences in convergence between convalutional network and transformer, all ResNet variants (ResNet-152, ResNet-101, and ResNet-50) undergo a uniform training period of 100 epochs. We apply a layer-wise decay rate of 0.55 for ViT-S, Swin-T, and ResNet-50; a decay rate of 0.4 for ViT-B, Swin-S, and ResNet-101; and a decay rate of 0.6 for ViT-L, Swin-B, and ResNet-152. The initial learning rates for fine-tuning are configured as follows: 0.002 for ViT-S, Swin-T, and ResNet-50; 0.0007 for ViT-B, Swin-S, and ResNet-101; and 0.0018 for ViT-L, Swin-B, and ResNet-152.

As is shown in Table. \ref{finetune}, our POA achieves a SOTA accuracy of $\textbf{85.3\%}$ on the ViT-L/16 backbone, and it demonstrates comparable accuracy when utilizing ViT-S/16 and ViT-B/16 backbones. We also report the fine-tuning results for the Swin and ResNet backbones in Table \ref{fine_swin_res}.

\begin{minipage}[c]{0.45\textwidth}
    \captionof{table}{Fine-tuning results on ImageNet-1K dataset.}
    \hspace{-2em}
    \setlength{\tabcolsep}{6pt}
    \begin{tabular}{cccc}
    \toprule
    Method  &Arch.    &Epo.  &Acc. \\
    \midrule
    DINO &ViT-S/16 &3200   &82.0 \\
    iBOT &ViT-S/16 &3200   &\textbf{82.3} \\
    \rowcolor{green!15}POA &ViT-S/16 &0   &82.1 \\
    \midrule
    BEiT &ViT-B/16 &800   &83.4 \\
    DINO &ViT-B/16 &1600   &83.6 \\
    iBOT &ViT-B/16 &1600   &\textbf{84.0} \\
    \rowcolor{green!15}POA &ViT-B/16 &0   &83.9 \\
    \midrule
    iBOT &ViT-L/16 &1200   &84.8 \\
    BEiT &ViT-L/16 &800   &85.2 \\
    \rowcolor{green!15}POA &ViT-L/16 &1200   &\textbf{85.3} \\
    \bottomrule
    \end{tabular}
    \label{finetune}
\end{minipage}
\begin{minipage}[c]{0.45\textwidth}
    \captionof{table}{Results of semi-supervised learning on ImageNet-1K. The $1\%$ and $10\%$ indicate the fractions of labeled data used. SD denotes self-distillation.}
    \setlength{\tabcolsep}{4pt}
    \begin{tabular}{cccc}
    \toprule
    Method  &Arch.    &$1\%$.  &$10\%$. \\
    \midrule
    SimCLRv2 &RN50 &57.9 &68.1 \\
    BYOL &RN50 &53.2 &68.8 \\
    SwAV &RN50 &53.9 &70.2 \\
    SimCLRv2 &\multirow{2}{*}{RN50} &\multirow{2}{*}{60.0} &\multirow{2}{*}{70.5} \\
    +SD & & &\\
    \rowcolor{green!15}POA &RN50 &\textbf{61.8}   &\textbf{73.1} \\
    \midrule
    DINO &ViT-S/16 &60.3 &74.3 \\
    iBOT &ViT-S/16 &61.9 &75.1 \\
    \rowcolor{green!15}POA &ViT-S/16 &\textbf{68.2}   &\textbf{75.9} \\
    \bottomrule
    \end{tabular}
    \label{semi}
\end{minipage}

\begin{table}[!ht]
    \caption{Fine-tuning results of Swin and ResNet backbone on ImageNet-1K dataset.}
    \begin{subtable}{0.45\linewidth}
    \setlength\tabcolsep{5pt}%
    \centering
    \begin{tabular}{lccc}
    \toprule
    Method     &Arac. &  Epo. & Acc\\
    \midrule
    POA & ResNet-50 &0  &77.8  \\
    POA & ResNet-101 &0 &80.0\\
    POA & ResNet-152 &2400 &81.1 \\
    \bottomrule
    \end{tabular}
    \caption{ResNet backbone.}
    \label{resnet_finetune}
    \end{subtable}
    \hspace{1em}
    \begin{subtable}{0.45\linewidth}
    \setlength\tabcolsep{8pt}%
    \centering
    \begin{tabular}{lccc}
    \toprule
    Method     &Arac. &  Epo. & Acc\\
    \midrule
    POA & Swin-T &0 &81.0  \\
    POA & Swin-S &0 &82.9 \\
    POA & Swin-B &1200 &83.7  \\
    \bottomrule
    \end{tabular}
    \caption{Swin backbone.}
    \label{swin_finetune}
    \end{subtable}
 \label{fine_swin_res}
\end{table}

\subsection{Semi-Supervised Learning Evaluation}
For semi-supervised learning, we concentrate our comparison on methods that adopt the unsupervised pre-training followed by supervised fine-tuning paradigm with partial labeled data. As shown in Table \ref{semi}, our method significantly outperforms iBOT when using only $1\%$ of labeled data, with an improvement of $\textbf{6.3\%}$. These results demonstrate our method's superior label efficiency. We attribute this performance enhancement primarily to the distillation loss $\mathcal{L}_{ES2}$ , which facilitates knowledge transfer from the intact model to its elastic counterpart. This effect mirrors the improvement observed in SimCLRv2, where self-distillation from a larger model contributes to performance gains.

\subsection{Unsupervised Learning Evaluation}
For evaluating the pre-trained model on unsupervised learning, we employ standard metrics such as accuracy (ACC), adjusted rand index (ARI), normalized mutual information (NMI), and the Fowlkes-Mallows index (FMI), following \cite{ibot}. We benchmark our POA with a ResNet-50 backbone against established methods like SimCLRv2 \cite{simclrv2}, Self-label \cite{self_label}, InfoMin \cite{info_min}, and SCAN \cite{scan}. Additionally, we compare POA with a ViT-S/16 backbone to DINO and iBOT. According to the results presented in Table. \ref{acc_unsupervised}, our POA method attains accuracies of \textbf{61.8\%} with ViT-S/16 and\textbf{ 55.7\%} with ResNet-50, respectively. These results indicate that the POA approach in self-supervised learning enables models to learn stronger visual semantical representation.

\begin{table}[!h]
  \caption{Unsupervised learning on ImageNet-1K dataset. "$\dagger$" denotes k-means clustering on frozen features extracted by backbones.}
  \label{acc_unsupervised}
  \centering
  \setlength{\tabcolsep}{13pt}
  \begin{tabular}{cccccc}
    \toprule
    Method& Arch. & ACC &ARI &NMI &FMI\\
    \midrule
    Self-label$^\dagger$& ResNet-50 &30.5 & 16.2 &75.4 & - \\
    InfoMin$^\dagger$& ResNet-50 &33.2 &14.7 & 68.8 & - \\
    SCAN & ResNet-50 &39.9 & 27.5 &72.0 & - \\
   \rowcolor{green!15} \textbf{ POA$^\dagger$} & ResNet-50 &\textbf{55.7} &\textbf{38.2}    &\textbf{79.9} &\textbf{38.9} \\
    \midrule
    DINO & ViT-S/16  & 41.4  & 29.8 & 76.8 & 32.8 \\
    iBOT &  ViT-S/16  &43.4 & 32.8 & 78.6 & 35.6 \\
    \rowcolor{green!15} \textbf{POA$^\dagger$} & ViT-S/16  &\textbf{61.8}   &\textbf{47.7}  &\textbf{82.5}  &\textbf{47.9} \\
    \bottomrule
  \end{tabular}

\end{table}

\begin{table}[!h]
    \caption{Transfer learning experiments by fine-tuning models pre-trained on various datasets. The Top-1 accuracy for the ViT-S/16 is presented on the left, and for the ViT-B/16 on the right.}
    \begin{subtable}{0.47\linewidth}
    \setlength\tabcolsep{1pt}%
    \centering
    \begin{tabular}{ccccccc}
    \toprule
    Method &Cif$_{10}$ &Cif$_{100}$ &iNa$_{18}$ &iNa$_{19}$ &Flwrs &Cars\\
    \midrule
    BEiT &98.6 &87.4 &68.5 &76.5 &96.4 &92.1 \\
    DINO &99.0 &90.5 &72.0 &78.2 &98.5 &93.0 \\
    iBOT &99.1 &90.7 &73.7 &78.5 &\textbf{98.6} &94.0 \\
    \rowcolor{green!15}POA &\textbf{99.1}  &\textbf{90.7}  &\textbf{74.2} &\textbf{79.1}    &98.4 &\textbf{94.2} \\
 
    \bottomrule
    \end{tabular}
    \end{subtable}
    \hspace{1em}
    \begin{subtable}{0.47\linewidth}
    \setlength\tabcolsep{1pt}%
    \centering
    \begin{tabular}{ccccccc}
    \toprule
    Method &Cif$_{10}$ &Cif$_{100}$ &iNa$_{18}$ &iNa$_{19}$ &Flwrs &Cars\\
    \midrule
    BEiT &99.0 &90.1 &72.3 &79.2 &98.0 &94.2\\
    DINO &99.1 &91.7 &72.6 &78.6 &98.8 &93.0\\
    iBOT &99.2 &92.2 &74.6 &79.6 &\textbf{98.9} &94.3\\
   \rowcolor{green!15}POA &\textbf{99.4} &\textbf{92.6}  &\textbf{76.2} &\textbf{81.7} &98.8 &\textbf{94.6}\\
    \bottomrule
    \end{tabular}
    \end{subtable}
 \label{trasfer_res}

\end{table}

\subsection{Transfer Learning} 
We evaluate transfer learning by pre-training models on ImageNet-1K and subsequently fine-tuning them on a variety of smaller datasets, adhering to the protocol established in \cite{vit}. The results are detailed in Table \ref{trasfer_res}. Our method achieves SOTA transfer performance compared to other self-supervised learning (SSL) approaches, with the exception of the Flowers dataset. Notably, we observe a more pronounced performance improvement over iBOT on larger datasets such as iNaturalist18 and iNaturalist19. This suggests that the results have not yet reached their peak, thereby providing a more effective measure for evaluating the quality of pre-trained features.

\subsection{k-NN Accuracies of Elastic Networks}
\paragraph{\textbf{Elastic Networks of ViT}} We present the k-NN evaluation accuracy of each elastic network derived from the pre-trained ViT trained by POA, as detailed in Table \ref{knn_acc_all_elastic_vit}, here $L_i$ and $D_i$ are the depths and widths of each elastic network, repectively.
\begin{table}[!ht]
  \caption{k-NN accuracy of elastic networks derived from pretrained ViT-L/16.}
  \label{knn_acc_all_elastic_vit}
  \centering
  \setlength{\tabcolsep}{3.8pt}
  \begin{tabular}{cccccccccccc}
    \toprule
    $L_i$/$D_i$& 384& 448& 512& 576& 640& 704& 768& 832& 896& 960& 1024 \\
    \midrule
    12&76.78& 78.42& 79.17& 79.78& 80.27& 80.68& 80.86& 81.02& 81.10& 81.14& 81.09\\
    13&76.84& 78.37& 79.17& 79.85& 80.83& 80.66& 80.93& 81.23& 81.24& 81.12& 81.15\\
    14&77.59& 78.95& 79.69& 80.26& 80.69& 81.04& 81.25& 81.36& 81.52& 81.60& 81.59\\
    15&77.89& 79.16& 79.86& 80.35& 80.75& 81.16& 81.40& 81.62& 81.62& 81.63& 81.58\\
    16&78.15& 79.30& 80.13& 80.75& 81.06& 81.36& 81.55& 81.78& 81.92& 81.77& 81.85\\
    17&78.34& 79.65& 80.26& 80.76& 81.11& 81.54& 81.63& 81.76& 82.02& 81.92& 81.88\\
    17&78.58& 79.75& 80.43& 80.92& 81.28& 81.69& 81.72& 81.99& 82.04& 82.02& 82.05\\
    19&78.62& 79.90& 80.52& 80.98& 81.32& 81.57& 81.83& 82.13& 82.09& 82.14& 82.12\\
    20&78.93& 79.97& 80.61& 81.17& 81.45& 81.79& 81.98& 82.17& 82.22& 82.15& 82.17\\
    21&78.99& 80.21& 80.73& 81.31& 81.49& 81.86& 82.05& 82.18& 82.32& 82.22& 82.20\\
    22&79.25& 80.25& 80.89& 81.36& 81.63& 81.87& 82.10& 82.31& 82.34& 82.36& 82.41\\
    23&79.37& 80.30& 80.89& 81.35& 81.73& 81.93& 82.19& 82.37& 82.41& 82.41& 82.39\\
    24&79.33& 80.28& 80.90& 81.33& 81.65& 81.90& 82.15& 82.35& 82.42& 82.28& 82.27\\
    \bottomrule
  \end{tabular}

\end{table}

\paragraph{\textbf{Elastic Networks of Swin}} For the Swin Transformer architecture, we designate Swin-T as the smallest elastic network and Swin-B as the largest. We explore elastic widths of 96, 112, and 128, with elastic depths varying from 12 to 24. The k-NN accuracies of total 39 elastic networks configuration are presented in Table. \ref{knn_acc_all_elastic_swin}.

\begin{table}[!ht]
  \caption{k-NN accuracy of elastic networks derived from pretrained Swin-B.}
  \label{knn_acc_all_elastic_swin}
  \centering
  \setlength{\tabcolsep}{1.8pt}
  \begin{tabular}{cccccccccccccc}
    \toprule
    $D_i$/$L_i$& 12& 13& 14& 15& 16& 17& 18& 19& 20& 21& 22 &23 &24 \\
    \midrule
    96 &77.48& 77.80& 78.07& 78.18& 78.17& 78.67& 78.81& 78.83& 79.05& 79.12& 79.20 &79.34 &79.31 \\
    112&77.84& 78.16& 78.33& 78.47& 78.47& 78.94& 79.03& 79.09& 79.22& 79.22& 79.32 &79.43 &79.43 \\
    128&77.90& 78.23& 78.47& 78.52& 78.52& 79.00& 79.07& 79.19& 79.45& 79.45& 79.53 &79.52 &79.63 \\
    \bottomrule
  \end{tabular}

\end{table}

\paragraph{\textbf{Elastic Network of ResNet}}
In the case of the ResNet architecture, we designate ResNet-50 as the smallest and ResNet-152 with wider middle layer in each building block as the largest elastic network configurations. It yields a total number of 465 distinct ResNet sub-networks with the combination of 3 widths and 155 depths configurations. For the sake of clarity, we present a subset of the k-NN accuracies of these elastic networks in Table \ref{knn_acc_all_elastic_resnet}. Here, $N_{2}$ and $N_{3}$ refer to the count of bottleneck building blocks in the second and third stages, respectively, while $W$ denotes the bottleneck dimension of middle layer in building block at the first stage.

\begin{table}[!h]

  \caption{k-NN accuracy of elastic networks derived from pretrained ResNet-152.}
  \label{knn_acc_all_elastic_resnet}
  \centering
  \setlength{\tabcolsep}{6.5pt}
  \begin{tabular}{ccccccccc}
    \toprule
    $W$/$N_{2}-N_{3}$& 4-6& 4-8& 4-10& 4-12& 4-14& 4-16& 4-16& 4-18 \\
    \midrule
    64 &73.44& 74.11& 74.32& 74.52& 74.92& 75.17& 75.19& 75.49 \\
    96 &74.41& 75.00& 75.18& 75.63& 75.95& 76.01& 76.11& 76.42 \\
    128&74.73& 75.43& 75.43& 75.81& 75.91& 76.20& 76.44& 76.47 \\
    \midrule
     $W$/$N_{2}-N_{3}$ &4-20 & 4-24& 8-26& 8-28& 8-30& 8-32& 8-34& 8-36\\
    \midrule
    64 &75.49& 75.77& 76.05& 76.20& 76.20& 76.33& 76.31& 76.38 \\
    96 &76.34& 76.63& 76.91& 76.95& 77.07& 77.13& 77.30& 77.14 \\
    128&76.53& 76.90& 77.13& 77.34& 77.31& 77.54& 77.59& 77.72 \\
    \bottomrule
  \end{tabular}
\end{table}

\subsection{Robustness Evaluation.}
\begin{figure}[!htp]
    \centering  
   \begin{subfigure}{0.47\linewidth} 
      \centering   
      \includegraphics[width=1\linewidth]{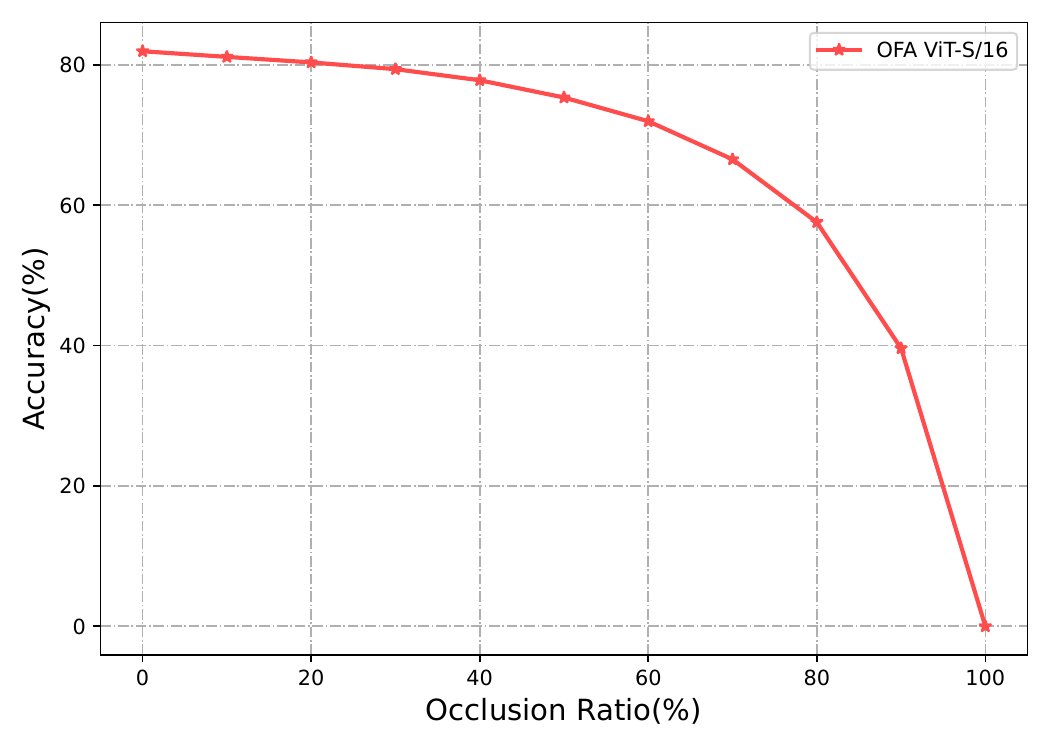}
        \caption{Robustness to occlusion with different ratio.}
        \label{fig:occlusion}
    \end{subfigure} 
    \begin{subfigure}{0.47\linewidth} 
      \centering   
      \includegraphics[width=1\linewidth]{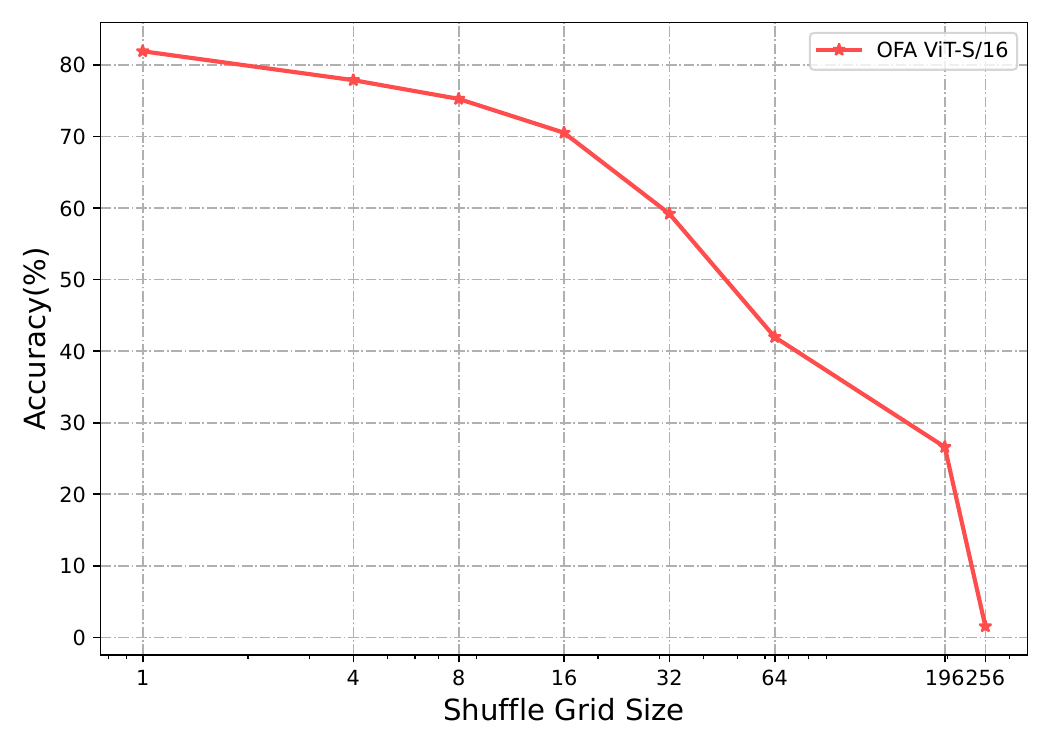}
        \caption{Robustness to shuffling with varying grid sizes.}
        \label{fig:shuffle}
    \end{subfigure} 
    \caption{Robustness to Occlusion and Shuffling.}
    \label{fig:robustness}
\end{figure}

\paragraph{\textbf{Robustness to Occlusion and Shuffling.}}
We evaluate the pre-trained model's robustness to occlusion and alterations in spatial structure by applying masking and shuffling to the input image patches. Detailed results for various occlusion ratios are depicted in Figure \ref{fig:occlusion}. Additionally, we present the effects of different shuffling grid sizes in Figure \ref{fig:shuffle}. 

\subsection{Additional Ablation Studies}
\paragraph{\textbf{Influence of Input Dimension Scaling Factor $\alpha_i$.}} To adapt the reduction of the input dimension, we apply a scaling factor $\alpha_i$ to weight parameters. We conduct comparative experiments to assess the impact of employing this scaling factor, with results presented in Table \ref{ablation_scaling}. The results indicate that the scaling factor $\alpha_i$ enhances the performance of an elastic network, particularly in cases where there is a significant reduction in width compared to the intact network, such as with ViT-S.

\begin{table}[!h] 
  \caption{Importance of Scaling Factor in POA SSL pre-training.}
  \label{ablation_scaling}
  \centering
  \setlength{\tabcolsep}{7pt}
  \begin{tabular}{ccccccc}
    \toprule
    \multirow{2}{*}{Scaling Factor} &   &k-NN & & &LP &  \\
    &ViT-S &ViT-B &ViT-L &ViT-S &ViT-B &ViT-L\\
    \midrule
    \checkmark &76.8 &80.9 &82.3 &77.6 &81.7 &83.6 \\
     &75.3 &80.8&82.3 &75.9&81.7&83.6  \\
    \bottomrule
  \end{tabular}
\end{table}

\paragraph{\textbf{Influence of Loss Weight $\lambda$.}} 
Within our POA framework, the parameter $\lambda$ regulates the balance between the loss contributions from the intact student and the elastic student. We assessed the performance impact of varying $\lambda$ during pre-training. The results presented in Table \ref{ablation_lambda} suggest that a larger value of $\lambda$, representing a greater loss weight for intact branch, enhances the performance of larger models like ViT-L. However, this same increase in $\lambda$ adversely affects the performance of the extracted smaller sub-networks such as ViT-S and ViT-B. Conversely, a smaller $\lambda$ value improves the performance of these smaller sub-networks while potentially diminishing the effectiveness of larger models. To achieve a more balanced outcome, we have chosen $\lambda = 0.8$ for our OPA approach.

\begin{table}[!h] 
  \caption{Influence of Loss Weight $\lambda$ in POA SSL pre-training.}
  \label{ablation_lambda}
  \centering
  \setlength{\tabcolsep}{10pt}
  \begin{tabular}{ccccccc}
    \toprule
    \multirow{2}{*}{$\lambda$ } &   &k-NN & & &LP &  \\
    &ViT-S &ViT-B &ViT-L &ViT-S &ViT-B &ViT-L\\
    \midrule
    0.6 &77.4 &81.0 &82.0 &78.3 &81.9 &83.1 \\
    0.7 &77.0 &80.9 &82.1 &77.9 &81.8 &83.4  \\
    0.8 &76.8 &80.9 &82.3 &77.6 &81.7 &83.6  \\
    0.9 &75.3 &80.0 &82.6 &76.5 &81.0 &84.0  \\
    \bottomrule
  \end{tabular}
\end{table}

\paragraph{\textbf{Influence of Probabilistic Sampling for Elastic Student.}} 
We provide the ablation study about probabilistic sampling for elastic student in Table. \ref{ablation_samlpling}. The result confirms our intuitive assumption that elastic networks with greater diversity should be sampled more frequently.

\begin{table}[!h] 
  \caption{Influence of sampling for elastic student in POA SSL pre-training.}
  \label{ablation_samlpling}
  \centering
  \setlength{\tabcolsep}{8pt}
  \begin{tabular}{ccccccc}
    \toprule
    \multirow{2}{*}{$S_w=S_d$ } &   &k-NN & & &LP &  \\
    &ViT-S &ViT-B &ViT-L &ViT-S &ViT-B &ViT-L\\
    \midrule
    1 &75.8 &80.6 &82.3 &76.5 &81.4 &83.6 \\
    2 &76.5 &80.7 &82.3 &77.4 &81.6 &83.6 \\
    3 &76.8 &80.9&82.3 &77.6&81.7 &83.6  \\
    \bottomrule
  \end{tabular}
\end{table}

\paragraph{\textbf{Influence of Number of Elastic Students.}} We investigate the impact of varying the number of candidate elastic networks in our POA. We manipulate the number of candidates by adjusting the sampling intervals of network widths and depths.  Except for the number of candidate elastic networks, all hyper-parameters and training settings remain constant. The comparative results are presented in Table \ref{ablation_num_cand}. From these results, we observe that the increase of the sampling interval, which effectively decreases the number of candidate networks, improves the k-NN accuracy for the derived ViT-S model. The primary reason is that with a constant number of iterations, a reduction in the total count of networks results in a higher proportion of smaller networks being sampled. This leads to more training iteration for these networks. However, this adjustment appears to have a negligible effect on the performance of ViT-L models, due to the existing of the intact student branch which is trained at each iteration.

\begin{table}[!h] 
  \caption{Influence of number of candidate elastic students in POA SSL pre-training.}
  \label{ablation_num_cand}
  \centering
  \setlength{\tabcolsep}{6pt}
  \begin{tabular}{ccccccc}
    \toprule
    Number of Candidates  & Interval of Elastic Sampling  &  &k-NN & \\
     ($\#$Widths $\times$ $\#$Depths) &(Width/Depth)  &ViT-S &ViT-B &ViT-L \\
    \midrule
    143($11\times13$) &64/1  &76.8 &80.9 &82.3 \\
    42($6\times7$) &128/2  &77.3 &80.9 &82.3 \\
    20($4\times5$) &192/3  &77.7 &80.8 &82.4 \\
    16($4\times4$) &192/4  &77.8 &81.1 &82.4 \\
    \bottomrule
  \end{tabular}
\end{table}

\subsection{Alternative Designs to Elastic Pre-training}

\begin{figure}[!h]
    \centering  
   \begin{subfigure}{0.95\linewidth} 
      \centering   
      \includegraphics[width=1\linewidth]{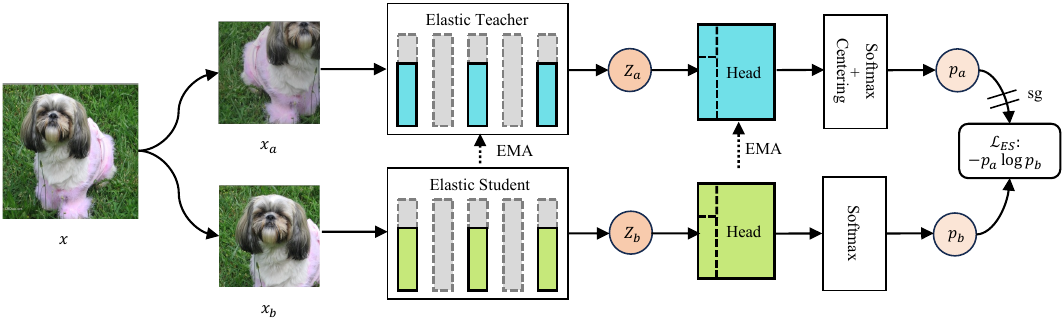}
        \caption{POA-V1: This variant of POA features both an elastic teacher and an elastic student, streamlining the architecture by ensuring both components are adaptable in size.}
        \label{fig:POA_variant1}
    \end{subfigure} 
    \begin{subfigure}{0.95\linewidth} 
      \centering   
      \includegraphics[width=1\linewidth]{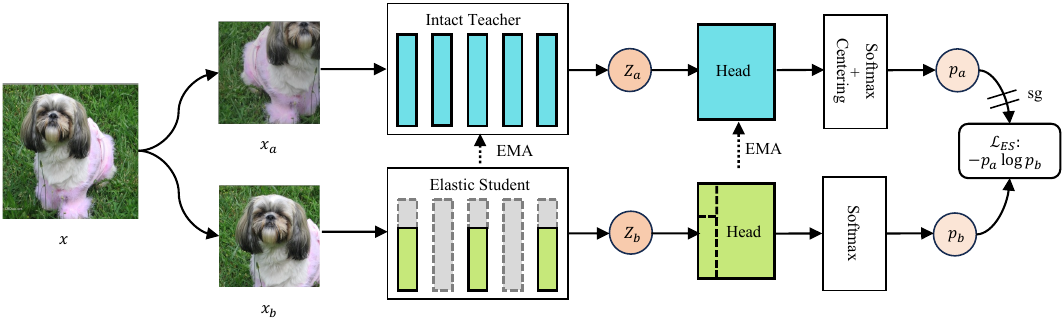}
        \caption{POA-V2: In this variant, POA includes an intact, intact teacher paired with an elastic student, allowing the smaller student model to learn from the larger, fully-trained teacher.}
        \label{fig:POA_variant2}
    \end{subfigure} 
    \begin{subfigure}{0.95\linewidth} 
      \centering   
      \includegraphics[width=1\linewidth]{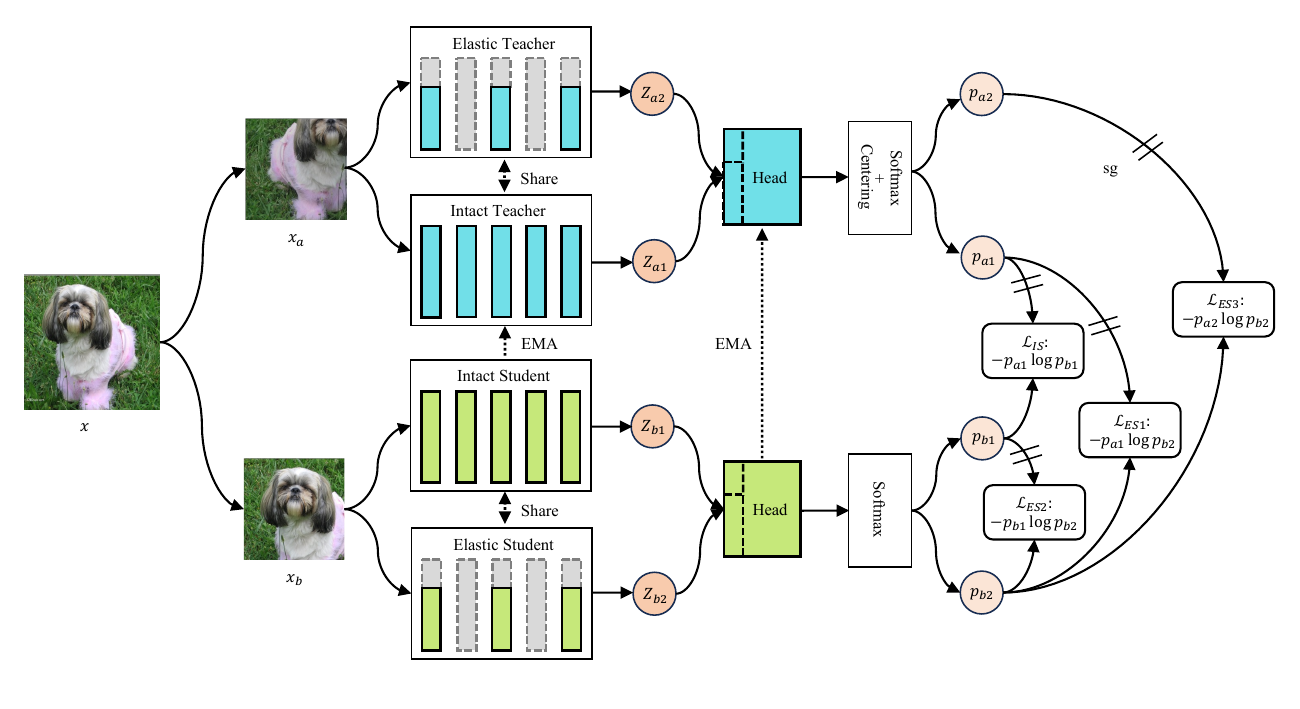}
        \caption{POA-V3: This variant of POA introduces an extra elastic teacher alongside the standard configuration, providing another potential for cross-view distillation within the framework.}
        \label{fig:POA_variant3}
    \end{subfigure} 
    \caption{Three alternative variants of POA.}
    \label{fig:cmp_variants}
\end{figure}

We provide a detailed illustration of the three variants mentioned in Sec.4.4 of our paper. In the first variant, illustrated in Figure \ref{fig:POA_variant1}, the intact student is removed from the POA framework and the teacher is adapted to be elastic, aligning with the architecture of the elastic student. The second variant, which also discards the intact student from POA, is depicted in Figure \ref{fig:POA_variant2}. The third variant introduces an additional elastic teacher branch that shares the architecture of the elastic student, facilitating cross-view distillation, shown in Figure \ref{fig:POA_variant3}.

\begin{figure}[!h]
    \centering  
      \centering   
      \includegraphics[width=1\linewidth]{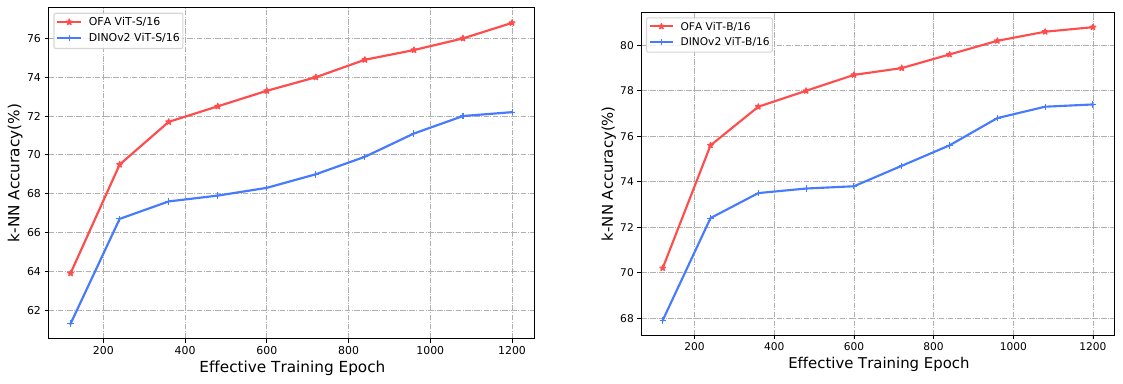}
    \caption{Comparison of k-NN accuracy during training.}
    \label{fig:cmp_converge}
\end{figure}

\subsection{Convergence Comparison with Single Pre-training Method}
By integrating same-view and cross-view distillation, POA achieves faster convergence speed, particularly for smaller-sized sub-networks. For instance, in the case of ViT-S, many existing self-supervised learning methods \cite{dino, ibot, mugs, ent} necessitate a substantial number of effective training epochs (3200 epochs) reach good performance. In contrast, the ViT-S model extracted from our POA framework, pre-trained for just 1200 epochs, outperforms those methods trained separately for 3200 epochs. Figure \ref{fig:cmp_converge} illustrates the k-NN evaluation accuracy progression throughout the 1200 effective epochs of training. It is evident that POA consistently delivers superior performance at each stage of the training process.

\begin{table}[!h] 
  \caption{Comparison of computational resources required.}
  \label{computing_resource}
  \centering
  \setlength{\tabcolsep}{6pt}
  \begin{tabular}{ccccccc}
    \toprule
    Method &Arch. &Epoch &Mem. &Batch Size &k-NN &GPU hours\\
    \midrule
    DINOv2 &ViT-L &1200 &41G &2048 &82.0 & 1152\\
    DINOv2 &ViT-L$\rightarrow$ViT-B &1200 &46G &4096  &79.7 & 1024\\
    DINOv2 &ViT-L$\rightarrow$ViT-S &1200 &35G &4096 &75.5 &928\\
    Total  & & & & & &3104 \\
    \midrule
    POA &ViT-L &1200 &77G &1600 &82.3 &2752 \\
    POA &ViT-B &0 &- &- &80.9 &0 \\
    POA &ViT-S &0 &- &- &76.8 &0 \\
    \bottomrule
  \end{tabular}
\end{table}

\subsection{Computational Resources Required}
Table. \ref{computing_resource} provides a detailed account of the computational resources required for training with a ViT backbone on 4 machines, each equipped with 8 A100 GPUs. We compare the time and GPU memory demands of our method to those of DINOv2, which incorporates self-supervised knowledge distillation \cite{seed} and yields superior performance compared to training the models independently. Notably, our approach can generate numerous elastic networks beyond the three primary 
structures: small, base, and large.

\section{Visualization}
\subsection{Visualization of Self-attention Map}
\begin{figure}[!h]
    \centering  
   \begin{subfigure}{0.495\linewidth} 
      \centering   
      \includegraphics[width=1\linewidth]{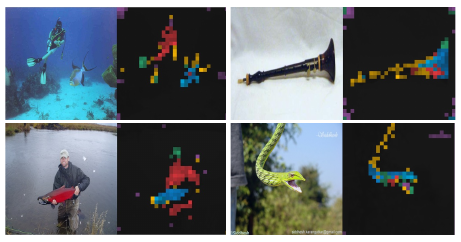}
        \caption{Self-attention map of POA.}
        \label{fig:att_POA}
    \end{subfigure} 
    \begin{subfigure}{0.495\linewidth} 
      \centering   
      \includegraphics[width=1\linewidth]{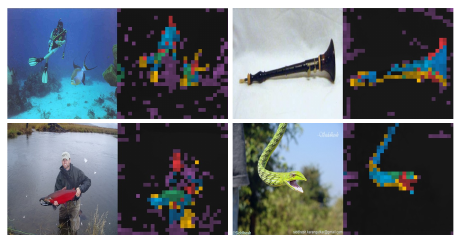}
        \caption{Self-attention map of DINOv2.}
        \label{fig:att_dinov2}
    \end{subfigure} 
    \caption{Visualization of Self-Attention Maps: we display the self-attention maps from multiple heads using distinct colors for differentiation. For both POA and DINOv2, we set the visualization threshold to 0.6, retaining top $60\%$ of the attention mass.}
    \label{fig:cmp_att}
\end{figure}

\begin{figure}[!h]
    \centering  
   \begin{subfigure}{0.91\linewidth} 
      \centering   
      \includegraphics[width=1\linewidth]{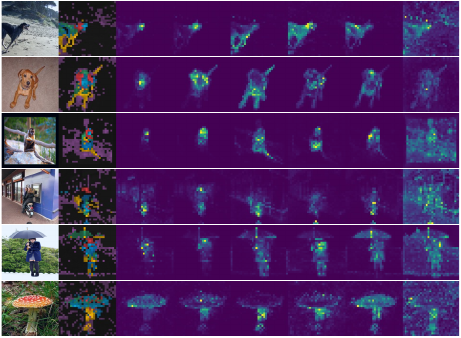}
        \caption{DIONv2}
        \label{fig:atten_dinov2}
    \end{subfigure} 
    \begin{subfigure}{0.91\linewidth} 
      \centering   
      \includegraphics[width=1\linewidth]{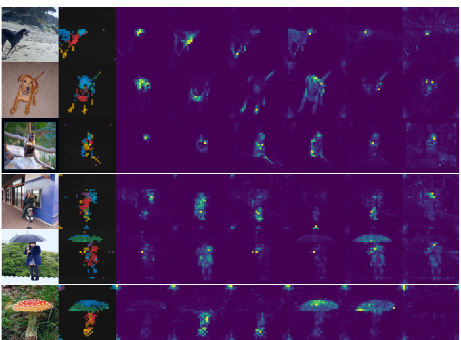}
        \caption{POA}
        \label{fig:atten_POA}
    \end{subfigure} 
    \caption{Visualization for self-attention map from multiple heads of the last layer in ViT-S/16. The results indicate that POA concentrates its attention more accurately on foreground objects than DINOv2 does.}
    \label{fig:cmp_att_more}
\end{figure}

We visualize the self-attention maps generated by the ViT-S/16 model, which is pre-trained using DINOv2 and our POA. For the visualizations, we select the class token as the query and represent attention maps from different heads of the final layer using distinct colors, as depicted in Figure \ref{fig:cmp_att}. The results indicate that POA's self-attention focuses more concentratedly on the foreground objects compared to DINOv2. For instance, in Figure \ref{fig:att_POA}, POA distinctly highlights the regions of interest associated with foreground elements (such as the human, fish, trumpet, and snake). In contrast, the DINOv2 generates attention maps exhibit a more dispersed focus, often including areas of the background. In Figure \ref{fig:cmp_att_more}, we showcase more self-attention map visualizations, comparing the outputs from multiple heads in the final layer of our method with those from DINOv2.

\subsection{Visualization of Correspondence}
\begin{figure}[!h]
    \centering  
   \begin{subfigure}{0.95\linewidth} 
      \centering   
      \includegraphics[width=1\linewidth]{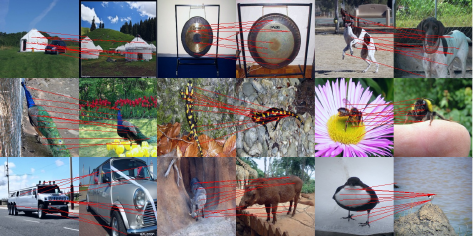}
        \caption{Correspondence between two different images of the same category}
    \end{subfigure} 
    \\
    \begin{subfigure}{0.95\linewidth} 
      \centering   
      \includegraphics[width=1\linewidth]{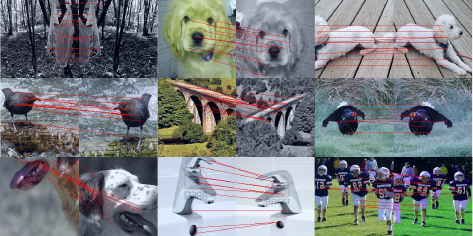}
        \caption{Correspondence between two different views of the same image}
    \end{subfigure} 
    \caption{Visualization of Correspondences: The top panel displays pairs of images sampled from two different views of a single image. The bottom panel shows pairs of images taken from two distinct images belonging to the same class.}
    \label{fig:correspondence}
\end{figure}

We conduct a correspondence task that involves matching overlapping patches from two different augmentations of the same image or patches from two distinct images labeled as the same class. We present visualizations of the these patches with the highest self-attention scores obtained from a ViT-S/16 model pre-trained by POA, averaging the scores across multiple heads in the final layer. Figure \ref{fig:correspondence} illustrates a selection of these image pair samples. The results indicate that POA excels in identifying correspondences both within varied views of a single image and across different segments of separate instances within the same class.

\subsection{Visualization of Pattern Layout for Class Token}

\begin{figure}[!h]
    \centering  
    \includegraphics[width=0.95\linewidth]{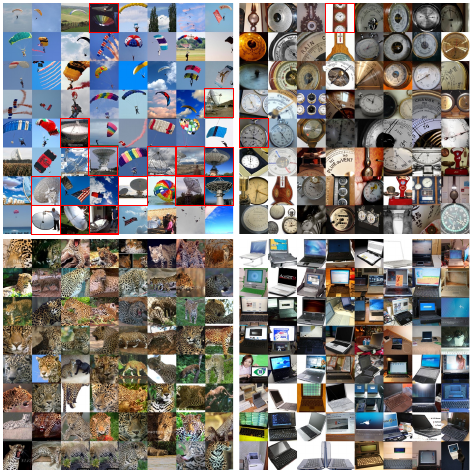}
    \caption{Visualization for pattern layout of class token of POA. We indicate that the prototypes effectively cluster images based on similar semantic features, even when they may span different categories.}
    \label{fig:layerout}
\end{figure}

Figure \ref{fig:layerout} presents a visualization of the pattern layout associated with the class token in ViT-S/16 trained by POA. We display images that have the top-64 similarity scores with each prototype in ImageNet validation set, arranged in an $8 \times 8$ grid. The results indicates the high-quality semantic structure achieved through the self-distillation process applied to cross-view images within our POA framework. Furthermore, we observe that the prototypes effectively cluster images based on similar semantic features, even when they may span different categories. For instance, in the top-left image of the grid, while the primary category is 'parachute', there are also images of related but distinct categories such as 'radio reflector' and 'umbrella', which are outlined in red boxes within these prototypes. Similarly, in the top-right image, the main category featured is 'odometer', but it also includes images of semantically similar objects like 'clock'.

%
%
\clearpage
\bibliographystyle{splncs04}
\bibliography{poa}
\end{document}